# Technical Report Statement

This document is a technical report consisting of two full-length works:

- Part I: *Graph Compression for Interpretable Graph Neural Network Inference At Scale* – the full version of our demonstration ExGIS.

- Part II: *Inference-friendly Graph Compression for Graph Neural Networks* – the full version of our manuscript under revision, VLDB.

These two parts are closely related and combined to provide a comprehensive reference for scalable, interpretable graph neural network inference.

# Graph Compression for Interpretable Graph Neural Network Inference At Scale

Yangxin Fan, Haolai Che, Mingjian Lu, Yinghui Wu
Case Western Reserve University
Cleveland, Ohio, USA
{yxf451,hxc859,mxl1171,yxw1650}@case.edu

## ABSTRACT

Graph Neural Networks (GNNs) have demonstrated promising performance in graph analysis. Nevertheless, the inference process of GNNs remains costly and hard to interpret, hindering their applications for large graphs. This paper introduces ExGIS, a parallel inference query engine to support explainable Graph Neural Network (GNNs) inference analysis in large graphs. (1) For a class of GNNs $\mathcal{M}^L$ with at most $L$ layers, and a graph $G$, ExGIS performs an offline, once-for-all compression of $G$ to a small graph $G_c$, such that for any inference query $Q$ that requests the output of any GNN $M \in \mathcal{M}^L$ on any node $v$ in $G$, $G_c$ can be directly queried to yield correct output without decompression. (2) Given a workload $W$ of inference queries that requests the output of GNNs from $\mathcal{M}$ over $G$, ExGIS perform fast online GNN inference and interpretation in parallel. It dynamically partitions $W$ to balance workloads, and (a) executes inference that only consults compressed graph $G_c$ without decompression, and (b) directly yields concise, explanatory subgraphs from $G_c$ that can clarify the query output with high fidelity, all in parallel. Moreover, ExGIS integrates visual, interactive interfaces for query performance analysis, and a Large Language Models (LLMs)- enabled interpreter to support user-friendly, natural language explanation of query outputs. Using real-world and synthetic large graphs, we experimentally verify the compression rate and scalability of ExGIS, and its application in interpretable anomaly detection over bitcoin transaction networks and academic networks. The source code, data, among other artifacts have been made available at https://github.com/nicej1899/ExGIS-Demo.

## 1 INTRODUCTION

Graph Neural Networks (GNNs) have shown promising performance in various analytical tasks. Despite their promising performances, GNNs incur expensive inference cost when $G$ is large [24]. The emerging need for large-scale, reliable data search, analysis and benchmarking require not only fast but also explainable inferences of GNNs where graphs play critical roles.

For example, money launderers who exploit Bitcoin blockchain transactions employ a variety of techniques to conceal illicit cryptocurrency activities and evade detection by law enforcement agencies and AI-based monitoring systems [17, 18]. While GNNs have been applied to detect suspicious accounts and activities in Bitcoin transaction networks, it is critical to make GNN inference scalable and interpretable to track them in fast growing transactions.

**Example 1:** Figure 1 illustrates a bitcoin transaction network $G$, where a node is an account IP addresses with attributes such as the number and amount of transactions. An edge $(v_1, v_2)$ means IP address $v_1$ committed a transaction to address $v_2$. Illicit accounts may utilize transaction patterns such as *Peel Chain* (transfering illicit assets along lengthy, numerous paths [17]) or *Spindle* ( multiple small cryptocurrency transactions [18]) to obscure illegal transactions.

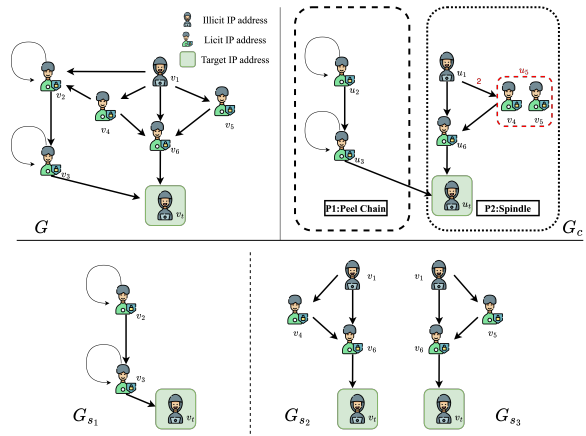

**Figure 1:** GNN-based anomaly detection in a bitcoin transaction network $G$. (1) ExGIS conducts an "once-for-all" compression to generate and distribute a compressed graph $G_c$ (top right) for online parallel inference query workload. (2) ExGIS processes fast online inference that detects "illicit" IPs by only consulting $G_c$, without decompression. (3) For a user designated account $v_t$ (in green box), its online, parallel explainer assembles three factual explanations that are well grounded by real-world strategies [17, 18].

Law enforcement agencies aim to uncover money laundry activities, making it essential to detect illicit accounts and generate explanations in large transaction networks. For $v_t$ as designated output nodes, we observe that the node $v_4$ and $v_5$ play a same "role" with same input features, and are "indistinguishable" for *any* inference query that involves a GNN with at most two layers and applies a same node update function (a GNN class $\mathcal{M}^2$) [1]. We can thus safely "merge" $v_4$ and $v_5$ and compress $G$ to a smaller counterpart $G_c$, and only perform inference over $G_c$ to infer $v_t$'s label.

A GNN inference process can be performed over a fragmented compressed graph $G_c$, *in parallel*, to detected "illicit" accounts. Better still, a GNN explainer can readily generate explanatory subgraphs as factual explanations [6] of the output of $v_t$, also in parallel over fragments of $G_c$, to assemble a set of diversified and comprehensive explanations $G_{s_1}$, $G_{s_2}$, $G_{s_3}$ to clarify "Why" $v_t$ is illicit, each grounded by a money laundry pattern, respectively. □

Several approaches have been separately developed to accelerate GNN inference by *e.g.,* simplifying model architecture, and to generate explanations for GNN output (see [22] for a survey).

These methods typically cope with specific classes of GNN, requires model internals (*e.g.*, parameters), or incurs considerable maintenance overhead for different GNN inference queries.

**ExGIS**. This paper introduces ExGIS, a parallel GNN inference and explanation framework to support large-scale and interpretable GNN inference. It has several unique features.

*Inference-friendly compression*. ExGIS adopts an *inference-friendly graph compression* scheme (IFGC) [1] to perform a "once-for-all" compression that compresses a large graph $G$ into a smaller counterpart $G_c$ with a memoization structure $\mathcal{T}$ that stores statistical neighborhood information of compressed nodes and edges. The compression scheme is provably guaranteed to preserve the output for *any* inference query for a GNN class that requests any GNN with layer $L$ and a node update function of the same form [1], via an efficient inference process directly over $G_c$ without decompression. This new feature is not addressed by prior GNN inference tools.

*Interpretable* GNN *inference at scale*. ExGIS effectively parallelizes a streamline of "inference-explanation" pipeline, which incorporates GNN inference and configurable explanation into an integrated online querying solution. Users only need to specify a graph, a set of designated test nodes $V_T$, and a GNN class, In one-click, it automatically creates, schedules, and distributes GNN inference and explanation joblets to assemble the inference output along with explanations. We show that ExGIS incurs a parallel cost that is only determined by the size of the compressed graph $G_c$ and the number of processors, regardless of how large $G$ is. While existing tools separate GNN inference and explanation as two independent task, ExGIS has made a first step towards a holistic "inference-and-explanation" workflow with scalability guarantee.

## 2 GRAPHS AND GRAPH NEURAL NETWORKS

**Graphs**. A directed graph $G = (V, E)$ has a set of nodes $V$ and a set of edges $E \subseteq V \times V$. Each node $v$ carries a tuple $T(v)$ of attributes and their values. The size of $G$, denoted as $|G|$, refers to the total number of its nodes and edges, *i.e.*, $|G| = |V| + |E|$.

**Graph Neural Networks**. A graph neural network (GNN) $\mathcal{M}$ is a mapping that takes as input a featurized representation $G = (X, A)$ to an output embedding matrix $Z$, *i.e.*, $\mathcal{M}(G) = Z$. Here $X$ is a matrix of node features, and $A$ is a (normalized) adjacency matrix of $G$. [1]

**Inference**. We take a query language perspective [3, 9] to characterize the inference process of GNNs. A GNN inference process is specified as a composition of *node update functions*.

*Node update function*. Given a GNN $\mathcal{M}$ with $L$ layers, a node update function $M_v$ uniformly computes the embedding of each node $v$ at each layer $k$ ($k \in [1, L]$), with a general recursive formula as

$$x_v^k = M_v^k(\Theta^k, \text{AGG}(X_u^{k-1}, x_v^{k-1}, \forall u \in N^k(v)))$$

which is specified by (1) the learned model parameters $\Theta^k$, (2) an aggregation function AGG (*e.g.*, $\sum$, CONCAT), and (3) the neighbors of $v$ that participate in the inference computation at the $k$-th layer (denoted $N^k(v)$). When $k=1$, $X_v^0 = X_v \in X$, *i.e.*, the input features.

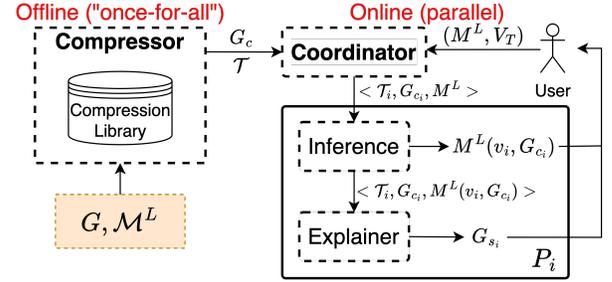

**Figure 2: ExGIS Workflow: Overview**

The *inference process* of a GNN $\mathcal{M}$ with $L$ layers takes as input a graph $G = (X, A)$, and computes the embedding $x_v^k$ for each node $v \in V$ at each layer $k \in [1, L]$, by recursively applying the node update function. A GNN $\mathcal{M}$ has a *fixed* inference process, if its node update function is specified by fixed input model parameters, layer number, and aggregator. It has a *deterministic* inference process, if $M(\cdot)$ always generates the same embedding for the same input.

We consider GNNs with fixed, deterministic inference processes. In practice, such GNNs are desired for consistent and robust performance. For simplicity, we assume that $M_v$ specifies a proper set of neighbors that participate the inference process as $N(v) \subseteq \{u | (u, v) \in E \text{ or } (v, u) \in E\}$. This allows us to include GNNs that exploits neighborhood sampling (such as GraphSAGE), and directed message passing into discussion. In general, inferences of representative GNNs are in PTIME [5, 24] (see Table 1).

**Classes of** GNNs. We say a set of fixed, deterministic GNNs $\mathbb{M}$ belongs to a *class of GNNs* $\mathbb{M}^L$, if for every GNN $\mathcal{M} \in \mathbb{M}$, (1) $\mathcal{M}$ has $L$ layers, and (2) $\mathcal{M}$ uses the same form of node update function $M_v^k$, for each node $v \in V$ and $k \in [1, L]$.

Table 1 summarizes several node update functions in their general forms for mainstream GNN classes. For example, Graph Convolution Networks (GCNs) [12] adopt a node update function as $X_v^k = \sigma(\Theta^k(\sum_{u \in N(v)} \frac{1}{\sqrt{d_u d_v}} x_u^{k-1}))$. Here $d_u$ or $d_v$ denotes the degree of node $u$ or $v$. $\sigma(.)$ is the non-linear activation function. A class of GNNs GCN[3] contains 3-layered GCNs that adopt such node update function. Note that two GNNs in the same class can have different $\Theta$ and output, given the same input.

## 3 SYSTEM OVERVIEW

**Workflow**. ExGIS works with a coordinator and a set of processors. It performs two major steps (see Fig. 2): an offline, "once-for-all" graph compression, and an online parallelized "Inference-and-Explain" pipeline upon the receiving or inference query workload.

*Offline Compression*. For an input graph $G$ and a class of GNN models $\mathbb{M}^L$ (see "Structural-preserving Compression"), the coordinator constructs a compression graph $G_c$ and a corresponding memoization table $\mathcal{T}$ (which collects useful auxiliary data) "once-for-all", to be consulted by any inference query $(M^L, v)$ that requests output $M^L(v, G)$ of any GNN $M^L$ in GNN class $\mathbb{M}^L$, for any node $v$ in $G$.

*Online Parallel "Inference-and-Explain"*. For each inference query $(M^L, v_i)$ from a query workload $(M^L, V_T)$, where $V_T$ is a set of nodes to be queried in $G$, an online "Inference-and-Explain" module consults the compressed graph $G_c$ and relevant fragments of memoization to conduct online parallel inference and explanation for each node $v_i \in V_T$, at each processor $P_j$ where $v_i$ resides. $P_j$ then registers inference output and the explanation graph to the coordinator in parallel to be eventually assembled and returned.

---

[1] A feature vector $X_v$ of a node $v$ can be a word embedding or one-hot encoding [8] of $T(v)$. $A$ is often normalized as $\hat{A} = A + I$, where $I$ is the identity matrix.

| GNNs Classes | Node Update Function (general form) | Training Cost | Inference Cost |
|---|---|---|---|
| Vanilla [19] | $X_v^k = \sigma(\Theta \cdot \text{AGG}(X_u^{k-1}, \forall u \in N(v)))$ | $O(L|E| + L|V|)$ | $O(L|E| + L|V|)$ |
| GCN [5, 12] | $X_v^k = \sigma(\Theta^k(\sum_{u \in N(v)} \frac{1}{\sqrt{d_u d_v}} x_u^{k-1}))$ | $O(L|E|F + L|V|F^2)$ | $O(L|E|F + L|V|F^2)$ |
| GAT [4, 20] | $X_v^k = \sigma(\sum_{u \in N(v)} \alpha_{uv} \Theta^k X_u^{k-1}))$ | $O(L|E|dF^2 + L|V|F^2)$ | $O(L|E|dF^2 + L|V|F^2)$ |
| GraphSAGE [5, 10] | $X_v^k = \sigma(\Theta^k \cdot (X_v^{k-1} || \text{AGG}(X_u^{k-1}, \forall u \in N(v))))$ | $O(L|V|dF + L|E|dF^2)$ | $O(L|V|dF + L|E|dF^2)$ |
| GIN [4, 21] | $X_v^k = \sigma(\text{MLP}((1 + \gamma)x_v^{k-1} + \sum_{u \in N(v)} x_u^{k-1})$ | $O(L|E|F + L|V|F^2)$ | $O(L|E|F + L|V|F^2)$ |

**Table 1: Comparison of Representative** GNNs **with node update functions, training cost, and inference cost.** $\sigma$: **an activation function** *e.g.,* **ReLU or LeakyReLU. AGG: aggregation function; can be** *e.g.,* **sum ($\sum$), average (Avg), or concatenation (||).** $L$, $|E|$, $|V|$, $F$, **and** $d$ **denote the number of layers, edges, nodes, features per node, and maximum node degree of** $G$, **respectively.**

**Inference-friendly Compression** [1]. A GNN class $\mathbb{M}^L$ refers to any set of GNNs with layers up to $L$ and adopt a node update function of the same form. Given a GNN model class $\mathbb{M}^L$ and a graph $G$, ExGIS induces a compressed graph $G_c$ by merges the nodes that are computationally "*indistinguishable*" based on structural and embedding equivalence (determined by a similarity threshold $\alpha$ of node attribute values). Meanwhile, ExGIS dynamically derives a memoization table $\mathcal{T}$ to "remember" neighborhood statistics (such as node degrees, edge attentions) of merged nodes as "scaling factors" that are needed to restore node embeddings from $G_c$.

ExGIS employs structural-preserving compression [1]. A compression scheme specifies a pair $(C, \mathcal{P})$. Compressor $C$ computes $G_c$ as the quotient graph of an equivalence relation $R^S$ over $G$, which is the non-empty, maximum equivalence relation that captures node pairs $(v, v')$ that are indistinguishable for inference queries for $\mathbb{M}^L$. $\mathcal{P}$ is an inference function that restores $\mathbb{M}^L([v], G_c)$ in $G_c$ (where the node $[v]$ is the equivalence class of $v$), with a matching scaling factor cached in $\mathcal{T}$. This ensures fast inference over $G_c$ without decompression.

**GNN Explainers**. ExGIS has a built-in library of post-hoc GNN explainers, which computes factual or counterfactual explanatory subgraphs for an output to be explained (*e.g.,* [6]). A factual explanatory subgraph for a node $v_i$ refers to a subgraph $G_S \subseteq G$ such that the GNN model $\mathbb{M}^L$ still preserves the same prediction for $v_i$, i.e., $\mathbb{M}^L(v_i, G_s) = \mathbb{M}^L(v_i, G)$. A counterfactual explanatory subgraph identifies a subgraph $G'_s \subseteq G$ such that the removal of $G'_s$ leads to the alteration of the prediction, i.e., $\mathbb{M}^L(v_i, G \setminus G'_s) \neq \mathbb{M}^L(v_i, G)$. These explanation subgraphs help identify the critical graph structures responsible for model predictions. For example, the three subgraphs $G_{s_1}, G_{s_2}, G_{s_3}$ in Fig. 1 are all factual explanation for the output of node $v_t$. Specially, $G_{s_1}$ is grounded by a peel chain pattern $P1$, and $G_{s_2}$ and $G_{s_3}$ witness spindle pattern $P2$. These explanations not only highlight influential illicit accounts to GNN's decision making, but also reveals their interactions for further inspection.

### 3.1 Inference-friendly Compression

We introduce the details of inference-friendly compression scheme (IFGC). Given a graph $G = (V, E)$, a *compressed graph* of $G$, denoted as $G_c = (V_c, E_c)$, is a graph where (1) each node $[v] \in V_c$ is a nonempty subset of $V$, and $V = \bigcup_{[v] \in V_c} [v]$; and (2) there is an edge $([v], [v']) \in E_c$, if there is a node $v \in [v]$ and $v' \in [v']$, such that $(v, v') \in E$. Note that $|V_c| \leq |V|$ and $|E_c| \leq |E|$. Hence, $|G_c| \leq |G|$.

**Inference-friendly Graph Compression.** Given a set of GNNs $\mathbb{M}$ and a graph $G$, an *inference-friendly graph compression*, denoted as IFGC, is a pair $(C, \mathcal{P})$ where

- $C$ is a compression function that computes a compressed graph $G_c$ of $G$ ($G_c = C(G)$);
- $\mathcal{P}$ is a function that restore the auxiliary information of nodes in $G_c$ to their counterparts in $G$; and moreover,
- $\mathcal{M}(G) = \mathcal{M}(\mathcal{P}(G_c))$, for *any* GNN $\mathcal{M} \in \mathbb{M}$.

An IFGC aims to generate a compressed graph $G_c$ with a smaller size, such that an inference query that requests output $\mathcal{M}(G, V_T)$ for any $V_T \subseteq V$ can be computed by a faster inference process of $\mathcal{M}$ over $G_c$ only, even with a query-time overhead incurred by $\mathcal{P}$.

**A Sufficient Condition.** We next introduce a sufficient condition for the existence of IFGC. To this end, we start with a notion of *inference-equivalent* relation. Given a class of GNN $\mathbb{M}^L$ and a graph $G$, a pair of nodes $(v, v')$ in $G$ are *inference equivalent w.r.t.* $\mathbb{M}^L$, denoted as $v \sim_M^L v'$, if for any $M \in \mathbb{M}^L$, $X_v^k = X_{v'}^k$ for any $k \in [1, L]$. One can readily infer that for any two nodes $v \sim_M^L v'$, $M(v, G) = M(v', G)$. That is, inference equivalence of nodes ensure that they are all "indistinguishable" for the inference of any GNN $\mathcal{M} \in \mathbb{M}^L$.

Denote the binary relation $(v, v')$ induced by inference equivalence as $R_M^L$, i.e., $(v, v') \in R_M^L$ if and only if $v \sim_M^L v'$. We say $R_M^L$ is *nontrivial* if there is at least one pair $(v, v') \in R_M^L$, where $v \neq v'$. We can readily verify the binary relation $R_M^L$ is an equivalence relation, i.e., it is reflexive, symmetric, and transitive.

The *equivalent class* of $v$ under an equivalence relation $R_M^L$, denoted as $[v]$, refers to the set $\{v' | (v, v') \in R_M^L\}$. The equivalent classes induced by the inference equivalence relation $R_M^L$ forms a node partition $V_R$ of $V$. The *quotient graph* induced by $R_M^L$ is a graph $G_R$ with nodes $V_R$ and edges $E_R$, where each node in $V_R$ is a distinct equivalent class induced by $R_M^L$, and there is an edge $([v], [v']) \in E_R$ if and only if there exists a node $v \in [v]$ and $v' \in [v']$, such that $(v, v') \in E$.

**Lemma 1:** *[1] Given a class of* GNNs $\mathbb{M}^L$ *and a graph* $G$, *a graph compression scheme* $(C, \mathcal{P})$ *is an IFGC w.r.t.* $\mathbb{M}^L$ *and* $G$, *if for any* $\mathcal{M} \in \mathbb{M}^L$, (1) $C(G)$ *computes a quotient graph* $G_c$ *induced by a nontrivial inference equivalent relation* $R_M^L$ *w.r.t.* $\mathbb{M}^L$ *and* $G$, *and* (2) $\mathcal{P}$ is a function that restores $X_v^k$ with $X_{[v]}^k$ by a scaling factor derived from auxiliary information of $v$, for each layer $k \in [1, L]$. $\square$

| GNNs | Node Update Function $M_v$ | equivalent rewriting $M_{[v]}$; scaling factors are marked in red | notes |
|---|---|---|---|
| Vanilla [19] | $X_v^k = \sigma(\Theta \cdot \text{AGG}(X_u^{k-1}, \forall u \in \mathcal{N}(v)))$ | $X_v^k = \sigma(\Theta \cdot \text{AGG}([\mathbf{v}]_{\mathbf{T}}(\mathbf{v}, [\mathbf{u}])X_{[u]}^{k-1} \mid \forall [u] \in \mathcal{N}([v])))$ | AGG: $\sum$ or AVG; for $AVG$, need to multiply by $RF_v$ |
| GCN [12] | $X_v^k = \sigma(\Theta^k(\sum_{u \in N(v)} \frac{1}{\sqrt{deg_u deg_v}} x_u^{k-1}))$ | $X_v^k = \sigma(\Theta^k(\sum_{[u] \in \mathcal{N}([v])} \frac{1}{\sqrt{deg_v}}[\mathbf{v}]_{\mathbf{T}}(\mathbf{v}, [\mathbf{u}])x_{[u]}^{k-1}))$ | $deg_v$: degree of node $v$ in $G$, topology sensitive |
| GAT [20] | $X_v^k = \sigma(\sum_{u \in N(v)} \alpha_{uv}\Theta^k X_u^{k-1})$ | $X_v^k = \sigma(\sum_{[u] \in \mathcal{N}([v])} [\mathbf{v}]_{\mathbf{T}}(\mathbf{v}, [\mathbf{u}])\Theta^k X_{[u]}^{k-1})$ | weight sensitive |
| GraphSAGE [10] | $X_v^k = \sigma(\Theta \cdot (X_v^{k-1}||\text{AGG}(X_u^{k-1}, \forall u \in \mathcal{N}(v))))$ | $X_v^k = \sigma(\Theta \cdot (X_{[v]}^{k-1}||\text{AGG}(RF_v \times [\mathbf{v}]_{\mathbf{T}}(\mathbf{v}, [\mathbf{u}])X_{[u]}^{k-1}, \forall [u] \in \mathcal{N}([v]))))$ | $||$: concatenation; AGG: AVG |
| GIN [21] | $X_v^k = \sigma(\text{MLP}((1+\gamma)x_v^{k-1} + \sum_{u \in N(v)} x_u^{k-1}))$ | $X_v^k = \sigma(\text{MLP}((1+\gamma)x_{[v]}^{k-1} + \sum_{[u] \in \mathcal{N}([v])} [\mathbf{v}]_{\mathbf{T}}(\mathbf{v}, [\mathbf{u}])x_{[u]}^{k-1}))$ | |

**Table 2: Rewriting of node update functions for mainstream GNN classes (scaling factors highlighted in red).**

The detailed proof is provided in [1]. We list examples of $\mathcal{P}$ and scaling factors for mainstream GNNs in Table 2.

We next introduce an IFGC for GNN inference, which specifies $R_M^L$ as an extended version of *structural equivalence*. The latter has origins in role equivalence in social science [15], and simulation equivalence of Kripke structures in model checking [2, 7]. By enforcing equivalence on embeddings and neighborhood connectivity, it ensures an IFGC to accelerate GNN inference *without decompression*.

*Structural equivalence.* Given a graph $G=(X, A)$, a *structural equivalence* relation, denoted as $R^S$, is a non-empty binary relation such that for any node pair $(v, v')$ in $G$, $(v, v') \in R^S$, if and only if:

- $X_v^0 = X_{v'}^0$, *i.e.,* $v$ and $v'$ have the same input features;
- for any neighbor $u$ of $v$ ($u \in N(v)$), there exists a neighbor $u'$ of $v'$ ($u'$ in $N(v')$), such that $(u, u') \in R^S$; and
- for any neighbor $u''$ of $v'$ in $N(v')$, there exists a neighbor $u'''$ of $v$ in $N(v)$, such that $(u'', u''') \in R^S$.

**Example 2:** Consider $G$ and $G_c$ in Fig. 1 and a GNN class with at most 2 graph convolutional layers that adopts a node update function that computes $v$'s embedding at $k$-th layer as $\sigma(\Theta^k(\sum_{u \in N(v)} \frac{1}{\sqrt{deg_u deg_v}} x_u^{k-1}))$. As $v_4$ and $v_5$ are merged into a node $u_5 = [v_4]$ in $G_c$, during inference, one just need to "recover" their original embeddings with a scaling factor corresponding to their original *in-degrees* at inference time. For example, for $v_4$, its embedding is directly rescaled as $\sigma(\Theta^k(\sum_{[u] \in N([v_4])} \frac{1}{\sqrt{deg}u_5} \mathcal{T}(\mathbf{v_4}, \mathbf{u_5})x_{u_5}^{k-1}))$, in constant time. $\square$

We next outline the compression algorithm and its corresponding inference algorithm without decompression.

**Compression Algorithm.** Given the input graph $G$, GNN model $M^L$, and a set of test nodes $V_T$, the compression algorithm, SPGC [1], generates the compressed graph $G_c$ and its corresponding memoization structure through the followings. First, SPGC computes *structural equivalence* relation $R^S$ of $G$. Next, SPGC (1) induces compressed graph $G_c$ as the *quotient graph* of $R^S$ and (2) caches the customized neighborhood statistics based on the choice of $M^L$ as the memoization structure $\mathcal{T}$ while constructing $G_c$. Finally, the compression algorithm returns the computed $G_c$ and $\mathcal{T}$ which can be applied to inference over any $V_T \subset V$ without decompression.

**Inference Algorithm.** Given $V_T$ and $G_c$, we outline an inference algorithm [1] that computes the inference results without decompression. For each node in $v \in V_T$, it utilizes the equivalent rewriting node update function $M_{[v]}$ as illustrated in Table 2 for mainstream GNNs to reweigh the embedding computation with the scaling factors directly looked up from $\mathcal{T}$. In this way, it computes the inference results $M^L(v, G_c)$ $\forall v \in V_T$ using the revised message-passing scheme that preserves inference accuracy over $V_T$.

**Example 3:** Consider $G$ in Fig. 1, a 2-layers GCN model $M^L$, and $V_T = \{v_t\}$, SPGC computes $R^S = \{(v_4, v_5)\}$ and induces compressed graph $G_c$ (shown in Fig. 1) where only $v_4$ and $v_5$ are merged into a supernode $u_5$ while other nodes remain uncompressed. SPGC constructs $\mathcal{T}$ while compressing $G$. The inference algorithm leverages $\mathcal{T}$ to compute inference result $M^L(v_t, G_c)$ of node $v_t$. $\square$

**Cost Analysis.** The compression cost is in $O(|G^L|)$ where $G^L$ is the subgraph induced by the $L$-hops neighbors of $V_T$. It takes $O(|G^L|)$ to compute $R^S$. It takes $O(|G^L|)$ to induce the compressed graph $G_c$ and compute $\mathcal{T}$ of $G^L$. As SPGC requires no decompression on neighborhood structures of nodes, an inference query can be directly applied to $G_c$ without incurring additional overhead. The inference algorithm is in $O(LF^2|V_T|)$ given that $M^L$ is a GCN model with at most $L$-layers and each node carries $F$ features.

## 3.2 GNN Explanation

We next describe the details of GNN explanation component. ExGIS, by default, adopts GVEX [6] as a representative GNN explainer. GVEX generates explanations for GNN predictions as *explanation views*. An explanation view is a two-tier structure that consists of (1) a set of graph patterns, which serve as human-interpretable motifs, and (2) a set of explanation subgraphs, which are substructures in input graphs that induce the prediction of a specific class label of interest. GVEX formalizes the generation of explanation views as an optimization problem that simultaneously maximizes *explainability* (*i.e.,* the aggregated influence of explanatory nodes) and *coverage* (*i.e.,* how well the explanation generalizes to include more important nodes and edges that are influential in the inference of a GNN).

To solve this optimization problem, GVEX introduces two algorithms with theoretical approximation guarantees. The main algorithm follows a "explain-and-summarize" strategy. First, it computes influence scores for each node based on their marginal effect on model prediction. To enable efficient influence computation, GVEX precomputes a Jacobian matrix $M_I$, where each entry encodes the sensitivity of model output with respect to input features. This matrix serves as a reusable backbone for evaluating node-level explainability in a once-for-all fashion. Then, it applies a diversified selection policy to collect a pool of candidate nodes with distinct and complementary explanatory power. Third, it invokes a structural verification module as a hard constraint: a candidate node is incorporated only if its inclusion into the current explanation subgraph satisfies factual and counterfactual properties. Upon verification, local explanation subgraphs are incrementally constructed around selected nodes. Finally, GVEX summarizes these subgraphs into frequent graph patterns and assembles an explanation view by selecting the patterns that jointly maximizes coverage and explainability. This greedy strategy guarantees a $\frac{1}{2}$-approximation for the overall view generation objective in terms of explainability and coverage, as a bi-criteria submodular function.

To further improved scalability and support large-scale or streaming graph settings, GVEX also introduces a second algorithm based on a streaming explanation paradigm. Instead of "materializing" all explanation subgraphs, this algorithm first summarize the graph data in a batched node stream to generate graph patterns (views) first. For each view, it incrementally identifies influential nodes, extracts local structures, and maintains the materialized explanation views under the same coverage constraint. By avoiding full enumeration, this approach significantly reduces memory and computation overhead, while still maintaining explanation quality. It provides an anytime $\frac{1}{4}$-approximation guarantee relative to the processed portion of the input, making it suitable for real-time or resource-constrained environments.

**Example 4:** Consider the compressed graph $G_C$ and the target node $v_t$ shown in Fig. 1. GVEX constructs an explanation view to clarify "why $v_t$ is classified as an illicit account?". The process starts by computing influence scores over nodes in the $L$-hop subgraph of $v_t$. Nodes with high but diverse explanatory impact—such as $v_2$, $v_4$, and $v_6$—are selected and verified for their factual or counterfactual properties. If a candidate passes verification, GVEX expands the explanation subgraph with it. For example, $G_{s1}$ captures a self-loop peel-chain flow from $v_2$ to $v_3$; $G_{s2}$ includes an upward path from $v_1$ to $v_4$ and $v_6$; and $G_{s3}$ forms a symmetric funnel into $v_t$. These subgraphs are later summarized into a compact explanation view composed of reusable graph patterns, offering a faithful and queryable rationale for the model's decision. □

**Cost Analysis** The computational cost of GVEX consists of three main parts: Jacobian matrix construction, explanation subgraph generation, and pattern summarization. It first incurs a one-time cost of $O(|V|^3)$ to compute the Jacobian matrix used for influence scoring. Subgraph construction involves at most $|V|$ rounds, with each round invoking GNN inference and structural validation. For a $k$-layer GNN, this costs $O(k \cdot C.\text{ul}(dD + D^2))$ per round, where $d$ is the average node degree, $D$ is the feature dimension, and $C.$ul is the upper bound on subgraph size. Summarizing $N$ verified subgraphs into patterns incurs $O(N \cdot T + N^2)$ time, where $T$ is the cost of subgraph matching. Overall, the total time cost of generating an explanation view for a graph is: $O(|V|^3 + k \cdot C.\text{ul}(dD + D^2) + N(N + T))$. In practice, $d$, $D$, $C.$ul, and $N$ are typically small due to the local and bounded nature of explanations.

---

**Algorithm 1 : Para-ExGIS**

1: **Input:** Compressed graph $G_c$, memoization structure $\mathcal{T}$, set of queried test nodes $V_T$, GNN model $M^L$, the coordinator $P_0$, set of parallel processors $\mathcal{P}$;
2: **Output:** Inference results and explanation graphs $\forall v \in V_T$.

3: **Set** $\mathcal{B} := \{b_i\}_{i=1}^{|\mathcal{P}|}$ where $\bigcup b_i = V_T$;
4: **for** $b_i \in \mathcal{B}$ **do**
5:    /* execute in parallel at $P_i \in \mathcal{P}$, $\forall v_{i,j} \in b_i$ */
6:    compute $\mathcal{T}_{i,j}$ from $\mathcal{T}$;
7:    induce $G_{c_{i,j}}$ from $G_c$;
8:    $J_{i,j} := < \mathcal{T}_{i,j}, G_{c_{i,j}}, M^L >$; /* inference joblet */
9:    $M^L(v_{i,j}, G_{c_{i,j}}) :=$ ParaInf $(J_{i,j})$;
10:    $\hat{J}_{i,j} := < \mathcal{T}_{i,j}, G_{c_{i,j}}, M^L(v_{i,j}, G_{c_{i,j}}) >$; /* explanation joblet */
11:    $G_{s_{i,j}} :=$ ParaExp $(\hat{J}_{i,j})$;
12: assemble inference results and explanations;
13: **return** $M^L(v_{i,j}, G_{c_{i,j}})$ and $G_{s_{i,j}}$ $\forall v_{i,j} \in V_T$;

1: **procedure** ParaInf$(J_{i,j})$
2:    $M^L$.forward$(G_{c_{i,j}})$ based on $\mathcal{T}_{i,j}$;
3:    compute inference result $M^L(\mathcal{T}_{i,j}, G_{c_{i,j}})$;
4:    **return** $M^L(\mathcal{T}_{i,j}, G_{c_{i,j}})$;

1: **procedure** ParaExp$(\hat{J}_{i,j})$
2:    compute explanation graph $G_{s_{i,j}}$;
3:    **return** $G_{s_{i,j}}$;

---

**Figure 3: Para-ExGIS: Parallel Inference and Explanation Implementation of ExGIS.**

## 4 PARALLEL "INFERENCE-AND-EXPLAIN"

In this section, we introduce the details of the parallel processing of the "inference-and-explain" phase. This component coordinates three major modules and algorithms that parallelize the online "inference-and-explain" pipeline, as illustrated in Fig. 4.

(1) **Coordinator**. Given the configuration tuple $(M^L, V_T)$ provided by the user, coordinator conduct the following steps: (1) retrieves the corresponding $G$ and $\mathcal{T}$ from the compressor; (2) generates parallel joblets $< \mathcal{T}_i, G_{c_i}, M^L >$ and sends them to processor $P_i$ for all $v_i \in V_T$. Here $M^L$ represents the GNNs model class with $L$ layers, specified by users from the representative GNN such as GCN, GAT, or GraphSAGE.

(2) **Compressor**. Given a graph $G$ and a class of GNNs $\mathcal{M}^L$, the compressor produces the compressed graph $G_c$ and the memoization structure $\mathcal{T}$. It performs offline "once-for-all" graph compression by leveraging a built-in library of compression methods, including our structural-preserving compression as well as the state-of-the-art graph compression approaches [13, 14].

(3) **Parallel Inference and Explainer**. We provide the pseudocode of Para-ExGIS (parallel inference and explanation algorithm) in Fig. 3. Given the joblet $< \mathcal{T}_i, G_{c_i}, M^L >$, the module executes the following two procedures in parallel (1) procedure ParaInf: computes the inference result $M^L(v_i, G_{c_i})$; and (2) procedure ParaExp: utilizes $< \mathcal{T}_i, G_{c_i}, M^L(v_i, G_{c_i}) >$ to generate explanation graph $G_{s_i}$, providing an interpretation of the inference for $v_i$.

**Coordinator**. The coordinator optimizes query workload with three components. (1) *Partitioner*: Upon receiving an inference query ($M^L$, $V_T$), it initializes a set of processors $\mathcal{P}$. At each $P_i \in \mathcal{P}$, it partitions $V_T$, the memoization table $\mathcal{T}$ to $\mathcal{T}_i$ and $\mathcal{T}_i$, and induces $G_{c_i}$ accordingly to $v_i \in V_T$. (2) *Load Balancer*: it assigns joblets to processors by distributing query workload among the processors, it minimizes the total time cost of inference and explanation on the entire set of $V_T$. (3) *Job Scheduler*: reschedules joblets $< \tau_i, G_{c_i}, M^L >$ based on estimated workload and current processor status, and allocate computational resources to dynamically rebalance the parallel "Inference-and-explain" computation.

*Pipelining*. At each processor $P_i$, given joblet $< \mathcal{T}_i, G_{c_i}, M^L >$, (1) it performs the inference for assigned $v_i \in V_T$ in parallel to generate output $M^L(v_i, G_{c_i})$. It next generates an explanatory task as a joblet $< \mathcal{T}_i, G_{c_i}, M^L(v_i, G_{c_i}) >$ and sends back to coordinator; (2) Upon receiving an assigned explanatory joblet $< \mathcal{T}_i, G_{c_i}, M^L(v_i, G_{c_i}) >$, it generates an explanation graph $G_{s_i}$ in parallel. The explainer constructs $G_{s_i}$ following the algorithm in [6] by default. The inference and explanation joblets follow a pipelined parallelism among all processes, to ensure that users receive analyzed output incrementally, rather than waiting or all inference queries to be evaluated.

**Para-ExGIS**. The Para-ExGIS algorithm, as illustrated in Fig. 3, takes as input $G_c$, $\mathcal{T}$, $V_T$, GNN model $M^L$, the coordinator $P_0$, and set of parallel processors $\mathcal{P}$. The partitioner at $P_0$ randomly partitions $V_T$ into a set of disjoint and equally sized node batches $\mathcal{B}$ (line 3). Para-ExGIS performs batch-level parallelism where the job manager at each $P_i \in \mathcal{P}$ coordinates the inference and explanation $\forall v_{i,j} \in b_i$ (line 4-11). The core functionality of Para-ExGIS consists of the followings: (1) $P_i$ computes the node-specific memoization structure $\mathcal{T}_{i,j}$ and induces its corresponding $G_{c_{i,j}}$; (2) the scheduler at $P_i$ schedules an inference joblet $J_{i,j}$ and invokes procedure ParaInf to compute the inference result; and (3) upon receiving the inference result, the scheduler schedules an explanation joblet $\hat{J}_{i,j}$ and invokes procedure ParaExp to compute explanation graph $G_{s_{i,j}}$. Para-ExGIS assembles the inference results and explanation graphs $\forall v \in V_T$ incrementally and returns them to the user (line 12-13).

*Parallel GNN Inference*. Upon receiving the inference joblet $J_{i,j}$ at processor $P_i$, Para-ExGIS invokes procedure ParaInf to implement the forward message passing with the revised node update function $M_{[v]}$ to compute inference result based on $G_{c_{i,j}}$ for node $v_{i,j} \in b_i$ where the scaling factors are looked up from $\mathcal{T}_{i,j}$ as illustrated in Sec. 3.1. ParaInf derives the inference result $M^L(v_{i,j}, G_{c_{i,j}})$ for $v_{i,j}$. At each parallel processor $P_i$, Para-ExGIS invokes ParaInf sequentially $|b_i|$ times to perform inference for each node $v_{i,j}$ in $b_i$.

*Parallel GNN Explanation*. Once receiving the explanation joblet $\hat{J}_{i,j}$, procedure ParaExp invokes the explanation algorithm described in Sec. 3.2 to generate the explanation graph $G_{s_{i,j}}$ for each node $v_{i,j} \in b_i$, a subgraph of $G_{c_{i,j}}$, computed based on the impacts of neighboring nodes on the inference result $M^L(v_{i,j}, G_{c_{i,j}})$ of $v_{i,j}$.

**Example 5:** As shown in Fig. 4, after obtaining $G_c$, $\mathcal{T}$, $M^L$, $V_T = \{a_1, a_2\}$, and the set of processors $\mathcal{P} = \{P_1, P_2\}$, Para-ExGIS partitions $V_T$ into $\mathcal{B} = \{\{a_1\}, \{a_2\}\}$. $P_1$ (resp. $P_2$) performs parallel inference and explanation for node $a_1$ (resp. $a_2$). We use $a_1$ as an example (the same steps apply to $a_2$). At $P_1$, Para-ExGIS computes $\mathcal{T}_1$

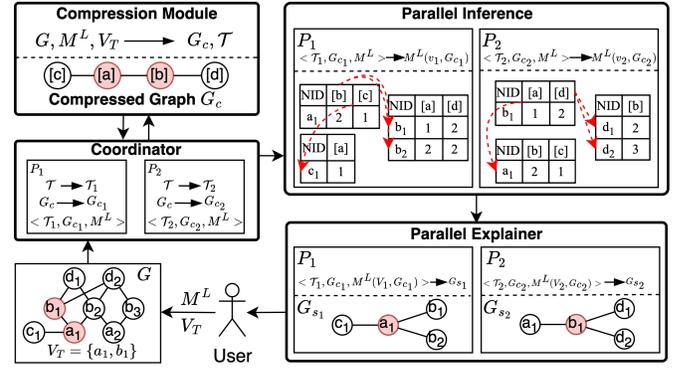

**Figure 4: Parallel Inference and Explainer with a running example:** $M^L \in$ GCN²; $V_T = \{a_1, b_1\}$; $\mathcal{T}_1$, $\mathcal{T}_2$: **partitioned subsets of** $\mathcal{T}$ **for** $a_1$ **and** $b_1$; $G_{s_1}$, $G_{s_2}$: **explanation graphs for** $a_1$ **and** $b_1$

and induces $G_{c_1}$. It creates an inference joblet $J_1 = < \mathcal{T}_1, G_{c_1}, M^L >$ and invokes procedure ParaInf to compute the inference result $M^L(a_1, G_{c_1})$. Using the inference result, Para-ExGIS creates an explanation joblet $\hat{J}_1 = < \mathcal{T}_1, G_{c_1}, M^L(a_1, G_{c_1}) >$ and invokes procedure ParaExp. Para-ExGIS computes the explanation graph $G_{s_1}$. Once both $P_1$ and $P_2$ are complete, Para-ExGIS returns the inference results and explanation graphs for $a_1$ and $a_2$ to the user. □

*Cost Analysis.* The total cost of Para-ExGIS is in $O((LF^2 + |G^L|) \frac{|V_T|}{n})$. $G^L$ is the subgraph induced by nodes within $L$ hops of the nodes in $V_T$. The parallel inference cost is in $O(\frac{LF^2|V_T|}{n})$ where $F$ is the number of node features and $n$ is the number of parallel processors. The inference cost of a single node is in $O(LF^2)$ as scaling factors can be retrieved from $\mathcal{T}$ in constant time. For $|V_T|$ queried nodes in total, ParaInf is in $O(\frac{LF^2|V_T|}{n})$. The parallel explanation cost is in $O(\frac{|G^L||V_T|}{n})$. Hence, the time cost of Para-ExGIS is in $O((LF^2 + |G^L|) \frac{|V_T|}{n})$. The time cost of Para-ExGIS is independent of the size of the original large graph $G$ and inverse proportional to $n$.

## 5 EXPERIMENTAL STUDY

**Experimental Setup.** We evaluate the performance of ExGIS and its parallel variant Para-ExGIS from two perspectives: inference accuracy and scalability. Our experiments are conducted on real-world and synthetic graphs using representative GNN models and compared against existing compressed inference baselines.

*Hardware specification.* All experiments are conducted on CWRU-managed HPC cluster. Each compute node used in our evaluation is equipped with an Intel Xeon Silver 4216 CPU at 2.10GHz (16 cores, 32 threads), 125GB of RAM, and a Tesla V100-SXM2 GPU with 32GB of memory. We use SLURM to allocate exclusive access to a single node per experiment. Parallel variants (e.g., Para-ExGIS) utilize up to 32 CPU threads and GPU acceleration when available. The CUDA version is 12.6, and the NVIDIA driver version is 560.35.

*Datasets.* We evaluate ExGIS on four real-world benchmarks and one synthetic large-scale graph. (1) **Cora** [16] is a citation graph where nodes are documents and edges represent citation links. Each node has a binary bag-of-words feature vector. (2) **Arxiv** [11] is another citation network built from arXiv papers, with each node feature being a 128-dimensional average of word embeddings from the paper's title and abstract. (3) **Yelp** [23] models user interactions in the Yelp platform, where nodes are users and edges represent friendships; node features are derived from review embeddings. (4) **Products** [11] is an Amazon co-purchase network, where nodes represent products and edges link items bought together. Node features are dense semantic vectors. All four datasets are used in our inference accuracy experiments.

For scalability evaluation, we use a synthetic graph, **BA-House**, with 2,020,0000 nodes and 12,055,7040 edges. It is constructed by attaching house motifs (5-node, 6-edge structures) to a Barabási-Albert backbone. Node features are 4-dimensional vectors indicating role positions (top, middle, base, or non-motif) within a motif. Across all datasets, we set the test set to 5% of the total nodes and assign targeted nodes for explanation generation accordingly.

*Graph Neural Networks.* We have pre-trained three classes of representative GNNs: GCNs[3] [12], GATs[3] [20], and GraphSAGE[3] [10], for each dataset. For a fair comparison, (1) for all the datasets, we consider node classification, and (2) all baseline methods are applied for the same set of GNNs.

*Baselines for comparison* We compare SPGC with two state-of-the-art compression methods: (1) DSpar [14], a graph sparsification method that performs edge down-sampling to preserve graph spectral information, and (2) FGC [13], a latest learning-based graph coarsening approach that learns a coarsened graph matrix and feature matrix to preserve desired graph properties, such as homophily. We are aware of other learning-based approaches, yet they are model-specific and require the learned model parameters. Our work is orthogonal to these methods, and is not directly comparable. For tests of DSpar and FGC, we set a longest waiting time limit as 5 hours. If the compression graph cannot be generated after 5 hours, we exclude such cases since the compression cost is already comparable to training a new GNNs from scratch.

*Evaluation Metrics.* Given a graph $G$, a GNN model class $\mathcal{M}^L$, and a Para-ExGIS operating over a compressed graph $G_c$, we assess system performance from both effectiveness and efficiency perspectives. (1) **Effectiveness.** We measure the model's predictive quality using accuracy and macro-averaged $F_1$-score over the test set. For the multi-class task on Yelp, we report micro $F_1$ to better reflect per-node classification quality. (2) **Efficiency.** We report the total inference time (in seconds) as the primary runtime cost, covering the full prediction pipeline. To capture the benefit of parallelization, we compute the per-node speedup as $\frac{t_{seq}}{t_{para}}$, where $t_{seq}$ and $t_{para}$ denote the average inference time per target node under sequential and parallel settings, respectively. All results are averaged over 200 inference trials for each GNN model and dataset.

**Inference Accuracy of ExGIS.** We evaluate the inference accuracy of ExGIS by comparing its backend compression algorithm SPGC with baseline methods DSpar and FGC across four real-world datasets: Cora, Arxiv, Yelp, and Products. Fig. 5(a) shows the accuracy and F1-score achieved by three GNN models (GCN, GAT, and GraphSAGE) when operating on the compressed graph $G_c$ produced by each method. Across all datasets and GNN variants,

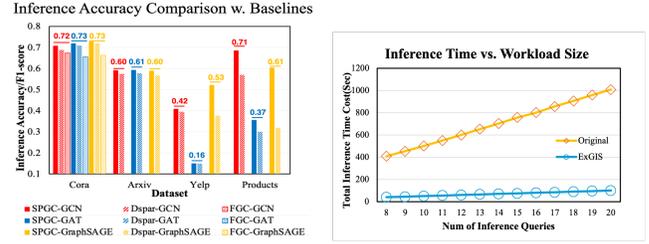

**(a)** ExGIS: **Accuracy on compressed graph $G_c$ across methods**

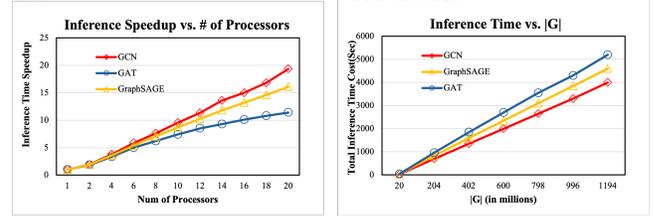

**(b) Para-ExGIS: Inference runtime vs. workload size (fixed #processors)**

**(c) Para-ExGIS: Inference runtime speedup vs. processor count (fixed workload)**

**(d) Para-ExGIS: Inference runtime vs. synthetic graph size (fixed #processors)**

**Figure 5: Para-ExGIS: Scalability and Accuracy.**

ExGIS consistently achieves competitive or superior accuracy. For example, on the *Yelp* dataset, where information loss due to compression is most severe, ExGIS still retains a significantly higher F1-score (e.g., 0.53 with GAT) compared to DSpar (0.42) and FGC (0.16). This demonstrates ExGIS's effectiveness in preserving model fidelity during compression. Importantly, the performance gap is more pronounced on large or structurally diverse graphs, where ExGIS's compression strategy better retains critical topological and feature-level information relevant to prediction.

**Impact of factors on Scalability of Para-ExGIS.**

*Varying Workload Size.* We evaluate the scalability of Para-ExGIS on the BA-House dataset by varying the number of independent inference queries (from 8 to 20), while fixing the number of processors to 20. Each query corresponds to a separate node classification task, either on the full original graph or on compressed $G_c$ generated by Para-ExGIS. In the original graph setting, inference is performed sequentially without parallel acceleration, resulting in a near-linear increase in total runtime as the number of queries grows—exceeding 1000 seconds at 20 queries. In contrast, Para-ExGIS leverages parallel inference on compressed graphs, keeping total runtime close to 100 seconds even under the largest workload. This demonstrates the scalability advantage and parallel efficiency of Para-ExGIS over conventional full-graph inference.

*Varying Processor Count.* We evaluate the parallel efficiency of Para-ExGIS by fixing the inference workload and varying the number of processors from 1 to 20 on the BA-House dataset. Experiments are conducted using three representative GNN architectures: GCN, GAT, and GraphSAGE. As shown in the results, all three models exhibit clear speed-up as the processor count increases, demonstrating the parallelizability of inference under Para-ExGIS. Among

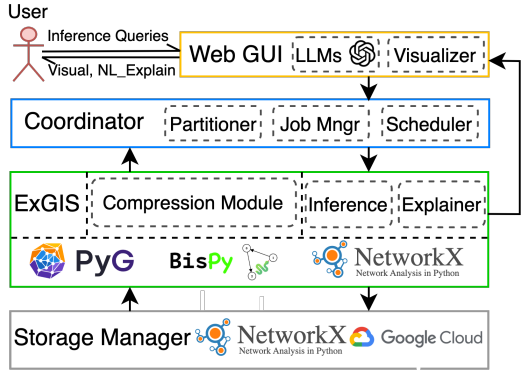

User
Inference Queries
Visual, NL_Explain

**Figure 6:** ExGIS **Architecture**

them, GCN achieves the highest speed-up (up to 19.34 with 20 processors), benefiting from its simple and layer-wise message aggregation which scales well in parallel execution. GraphSAGE follows with moderate performance due to its localized sampling and moderate computation per node. In contrast, GAT achieves the lowest speed-up due to the overhead introduced by attention mechanisms, which involve additional computation and memory access patterns that are less efficient to parallelize. Nevertheless, all models significantly reduce inference time under Para-ExGIS, confirming its general effectiveness across GNN backbones.

*Varying Synthetic Graph Size.* To evaluate the scalability of Para-ExGIS with respect to the input graph size, we extend beyond the baseline 20 million-node synthetic BA-House graph and construct larger graphs by scaling the generator parameters. While fixing the number of processors and keeping other variables constant (e.g., number of inference queries, model architecture), we observe the total inference time under each graph size. The results show that the inference runtime grows approximately linearly with graph size across all tested models (GCN, GAT, and GraphSAGE), validating the scalability of our framework. Notably, Para-ExGIS sustains reasonable inference latency even for graphs with over one billion nodes, demonstrating its practical applicability to large-scale real-world deployments.

## 6 SYSTEM DEPLOYMENT AND ARCHITECTURE

The multi-tier ExGIS architecture has been deployed at HPC cluster at CWRU supported on well established data systems, as illustrated in Fig. 6. (1) The *Web GUI* consists of (a) portals to collect metadata of configuration and receiving inference query workload; (b) LLM interfaces for NL interpretations; and (c) a visualizer to support visual analysis of compressed graphs, explanation graphs, and inference-time performance evaluation. (2) The *Coordinator* hosts the job scheduler to i) manage the allocations of computational resources; and ii) coordinate message exchanging among the compression module, explainer, GUI, and storage manager. (3) The *ExGIS Core* contains critical computational modules: offline Compression and online parallel Inference and Explainer, supported by built-in libraries *e.g.,* PyTorch Geometric, BisPy, NetworkX and LLMs including Llama 3.3 and GPT-4o. (4) The *Storage Layer* hosts all graphs as NetworkX objects. The other artifacts including GNNs models and metadata are stored as JSON objects.

## 7 DEMONSTRATION SCENARIO

**Datasets**. We demonstrate ExGIS with Elliptic Bitcoin transactions network (203,769 nodes, 234,355 edges), and Cora citation network (2,708 nodes, 5,429 edges). We report complete tests with two graph compression baseline methods across six diverse datasets [1].

**"Inference & Explain" in One-Click**. We start with a walkthrough of ExGIS. Users begin by selecting datasets, configuring datasets, compression parameters (compression scheme and parameters), mainstream GNN classes (*e.g.*, GCN, GAT, or GraphSAGE) with layer number, test nodes, and number of processors. With a single click on "Run Compression & Inference," the system automates offline compression to cold start, streamlines the workload optimization for parallel execution. Users can then request ad-hoc inference queries by simply choosing nodes in the compressed graph panel, and readily browse and inspect the visual output and explanations.

**Accuracy & Scalability Analysis**. Our demonstration highlights ExGIS's performance metrics. The Statistics Analysis tab shows that for Bitcoin transaction networks, ExGIS achieves compression ratios of 92.4% for nodes and 91.3% for edges while maintaining 100% inference accuracy with a 2-layer GCN. The time statistics reveal excellent efficiency: compression (0.039s), inference (0.003s), and explanation generation (0.007s). As shown in Fig. 7, ExGIS maintains competitive accuracy compared to baselines across datasets while offering superior scalability - execution time remains nearly constant for SPGC as queries increase, unlike the linear growth observed with original graphs. Due to limited space, we report more scalability tests and analysis in [1].

**Interpretable Network Anomaly Detection**. We invite users to experience user cases of ExGIS in two applications.

*Illicit IP Account Detection.* Using Elliptic Bitcoin transaction network, users can specify inference queries over designated IP account and visually inspect most influential illicit and licit accounts and transactions to GNN output, observe the impact of different distribution of labels to the decision making of GNNs, and be informed by natural language alerts on suspicious activities.

*Topic Analysis in Citation Networks.* Our second scenario uses the Cora academic citation network. Users can conveniently choose papers of interests and test GNNs of choices to infer potential topics. The explanation reveals influential citations that may influence GNN inference for the particular topic via a subnetwork of citations, hence in turn suggesting useful literature and collaboration.

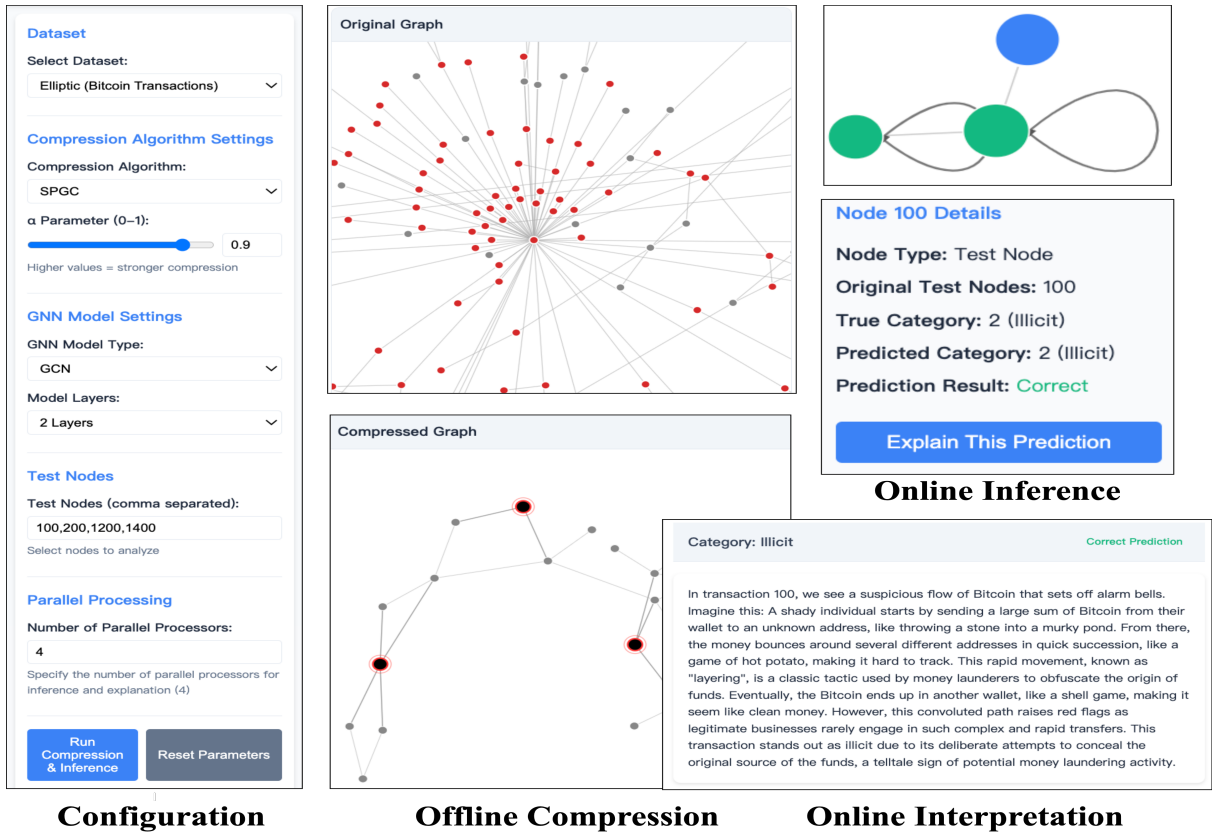

**Figure 7: Visual Interfaces of** ExGIS

# Inference-friendly Graph Compression for Graph Neural Networks

Yangxin Fan, Haolai Che, Yinghui Wu

Case Western Reserve University

Cleveland, Ohio, USA

{yxf451,hxc859,yxw1650}@case.edu

## ABSTRACT

Graph Neural Networks (GNNs) have demonstrated promising performance in graph analysis. Nevertheless, the inference process of GNNs remains costly, hindering their applications for large graphs. This paper proposes *inference-friendly graph compression* (IFGC), a graph compression scheme to accelerate GNNs inference. Given a graph $G$ and a GNN $M$, an IFGC computes a small compressed graph $G_c$, to best preserve the inference result of $M$ over $G$, such that the result can be directly inferred by accessing $G_c$ with no or little decompression cost. (1) We characterize IFGC with a class of inference equivalence relation. The relation captures the node pairs in $G$ that are not distinguishable for GNN inference. (2) We introduce three practical specifications of IFGC for representative GNNs: structural preserving compression (SPGC), which computes $G_c$ that can be directly processed by GNN inference without decompression; $(\alpha, r)$-compression, that allows for a configurable trade-off between compression ratio and inference quality, and anchored compression that preserves inference results for specific nodes of interest. For each scheme, we introduce compression and inference algorithms with guarantees of efficiency and quality of the inferred results. We conduct extensive experiments on diverse sets of large-scale graphs, which verifies the effectiveness and efficiency of our graph compression approaches.

## 1 INTRODUCTION

Graph Neural Networks (GNNs) have shown promising performance in various analytical tasks such as node classification [31] and link prediction [55]. In general, a GNN $M$ converts an input graph $G$ (as a pair $(X, A)$ of node feature matrix $X$ and an adjacency matrix $A$), to a vector representation ("embeddings") $M(G)$ via multiple layers. For each node, each layer applies a same "node update function" to uniformly update its embedding as a weighted aggregation of embeddings from its neighbors, subsequently transforming it towards an output embedding. The training of $M$ is to optimize its model parameters ("weights") and obtain a proper update function to make it best fits a set of training data in a training graph. Given an input (test) graph $G$, the *inference* of $M$ applies the node update function to generate output embeddings $M(G)$ (a matrix of node embeddings). $M(G)$ can be post-processed to task-specific output, such as class labels for node classification.

Despite their promising performances, GNNs incur expensive inference process when $G$ is large [13, 53, 58]. The emerging need for large-scale testing, fine-tuning and benchmarking of graph learning models, require fast inferences of GNNs under various configurations. Consider the following scenarios.

(1) *Inference in large networks*. For a graph $G$ with $|V|$ nodes and $|E|$ edges (where $V$ and $E$ refers to its node and edge set), with on average $F$ features per node, an $L$-layered GNN $M$ may typically take $O(L|E|dF^2 + L|V|F^2)$ time [13] (as summarized in Table 3). This can be prohibitively expensive for large real-world graphs.

(2) *Real-time Inference*. GNN models have been developed for *e.g.*, traffic analysis [42], social recommendation [18], forecasting [2, 19, 30], computer vision [46], and edge devices [58]. Such scenarios often require real-time response at *e.g.*, milliseconds [58]. In such cases, even a linear time inference of "small" GNNs (when $F$ and $L$ are small constants) may still not be feasible for large graphs (when $|V|$ and $|E|$ are large).

(3) *Fine-tuning & Benchmarking*. Fine-tuning and testing pre-trained GNNs to adapt them for various domain-specific tasks is a routine process in GNN-based data analysis for *e.g.*, materials sciences, biomedicine, social science, and geosciences [25, 43, 48, 57]. Inference tests of large pool of "candidate" GNNs over domain-specific graph data (such as knowledge graphs) is a cornerstone in such context. Fast GNN inference can accelerate large-scale domain-specific testing and benchmarking.

Several approaches have been developed to accelerate GNN inference, by simplifying model architecture [41], (learning) to optimize inference process [47], or data sampling [14]. These methods typically works with specific GNN $M$, requires prior knowledge of its internals (*e.g.*, model parameter values), and may incur new computation overhead each time a different GNN $M$ is specified.

**Compressing graphs for** GNN inference. Unlike prior "model-specific" approaches, we propose a model-agnostic, "*once-for-all*" graph compression scheme to accelerate GNN inference, for a *set* of GNNs. Consider a set of GNNs $\mathbb{M}$ (a GNN "class") with the same form of inference, which apply the same "type" of node update function but only differs in model weights (see Example 1). The inference of a GNN $M$ over $G$ can be characterized as an "inference query" [6, 23], which invokes the inference of $M$ to compute $M(G)$.

We advocate an "*inference-friendly*" graph compression scheme for GNN inference at scale. Given $\mathbb{M}$ and a large graph $G$,

- It uses a compression function $C$ to compute a smaller counterpart $G_c$ of $G$ "once-for-all", for any GNN $M \in \mathbb{M}$;
- For any inference query that requests $M(G)$ for a specific GNN $M \in \mathbb{M}$, it performs an inference directly over $G_c$ instead of $G$ to compute $M(G_c)$, with a reduced time cost, such that $M(G_c)$ (approximately) equals $M(G)$.

Such a compression is desirable: (1) It readily reduces the cost for any single inference query that computes $M(G)$; (2) An inference query often does not require the entire output $M(G)$ but only a fraction $M(G, V_T) \subseteq M(G)$ of a specified test (node) set $V_T$ of interests; (3) Multiple inference queries can be posed to request output from different GNNs in $\mathbb{M}$ in $G$. For any workload with

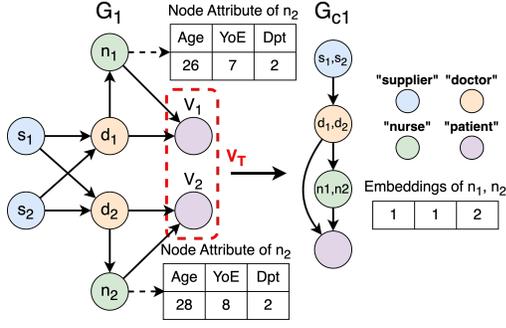

**(a) Social role classification: a hospital network ($G_1$) can be "compressed" by merging nodes with equivalent social roles for testing a GCN-based classifier (adapted from [11], node attributes include Age, YoE (years of experience), and Dpt (department)).**

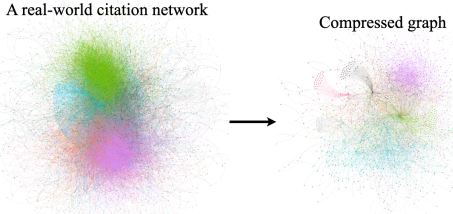

**(b) Visualization of a fraction of real-world citation network [28] ($|G| = 1,335,586$) and its compressed counterpart ($|G_c| = 148,887$) for a GraphSAGE-based node classification. 88.6% of nodes and edges are compressed, reducing inference cost by 92.0%, achieving 12.5 times speed-up with up to 6.7% loss of accuracy.**

**Figure 1: Compression Scheme to scale node classification.**

inference queries that specify any GNN $\mathcal{M} \in \mathbb{M}$ and any $V_T$ from $G$, one only need to compute $G_c$ once, to reduce the total inference cost of the workload. These benefits applications in large-scale tests over large graphs, real-time inference and benchmarking, as aforementioned.

While desirable, *is such a compression scheme doable?* We illustrate a case in the following example.

**Example 1:** Consider a 3-layer Vanilla GNN $\mathcal{M}$ [45] as a node classifier that assigns role labels {supplier, doctor, nurse, patient} in a social healthcare network $G_1$ (illustrated in Fig. 1). Each node in $G_1$ has attributes such as role, age group, department, etc. To infer the roles of test nodes $V_T = \{v_1, v_2\}$, an inference process of $\mathcal{M}$ starts by propagating a node feature matrix $X$ with a node update function $M_v$. Via a 3-layer forward message passing, the embeddings of $v_1$ and $v_2$ are obtained, quantifying the likelihood of them being assigned to one of the labels. As the probability of "patient" is the highest for both, it infers both labels as "patient".

We take a closer look at the update function $M_v$ at layer $k$:

$$X_v^k = \sigma\left(\Theta \cdot \sum_{u \in N(v)} X_u^{k-1}\right)$$

where $X_v^{k-1}$ (resp. $X_v^k$) is the embedding of a node $v$ at the $(k-1)$-th (resp. $k$-th) layer; $\sigma$ is an activation function, $N(v)$ refers to the neighbors of node $v$, and $\Theta$ refers to the learned weight matrix (same across all layers in the GNN).

If the input features of a pair of nodes $x_1$ and $x_2$ are the same (with $x$ ranges over $s$, $d$, $n$ and $v$), then the embedding of $x_1$ and $x_2$ will be the same during the inference computation, as long as the above the node update function is applied for any fixed model weights $\Theta$ and any fixed total number of layers. That is, $x_1$ and $x_2$ are "indistinguishable" for the inference of any 3-layered GNN $\mathcal{M}$ that adopts the above node update function $M$ with the same aggregator AGG, regardless of how its $\Theta$ changes.

Note that feature equivalence does not necessarily mean that $x_1$ and $x_2$ have exactly the same attribute values as input. For example, while $n_1$ and $n_2$ refer to a 26 years old and a 28 years old nurse, respectively, their ages, years of experience (YoE), and department (Dpt) fall in the same group via (categorical or one-hot) encoding. Hence, they have the same input feature.

By "recursively" merging all such node pairs into a "group node" that are connected to neighbors that are also indistinguishable groups (e.g., $[s] = \{s_1, s_2\}$, $[d] = \{d_1, d_2\}$, $[n] = \{n_1, n_2\}$, and $[v] = \{v_1, v_2\}$), a smaller graph $G_c$ can be obtained. An inference directly over $G_c$ can yield the same output for the test nodes $v_1$ and $v_2$ for $\mathcal{M}$ without decompression. To see this, one just need to "recover" their original value with a constant factor 2 (their original degrees) as auxiliary information at query time, for each layer:

$$X_{v_1}^k = X_{v_2}^k = 2 \times X_{[v]}^k = 2 \times \sigma(\Theta \cdot X_{[n]}^{k-1})$$

Such aggregated neighborhood information (e.g., degrees, edge weights/attentions, or hyper-parameters) can be readily "remembered" at compression time, and be retrieved in constant time. This indicates an overall cheaper inference cost, and an exact query-time restore of the original embedding, for any node in $G$.

Better still, we only need to compute $G_c$ "once-for-all", to reduce the unnecessary inference cost for any set of inference queries that specify a 3-layer GNN $\mathcal{M}$ (regardless of their weights $\Theta$) that uses the same node update function $M$, and for any $V_T$ in $G$. □

The above example verifies the possibility of a graph compression scheme by finding and merging node pairs that are indistinguishable for the inference process of GNNs. Our study verifies that real-world graphs are indeed highly compressible with such structures, and if compressed, well preserve inference output with no or small loss of accuracy, and meanwhile significantly reduce unnecessary inference computation (see Fig. 1 (b)).

**Contributions**. Our main contributions are as follows.

(1) We formally introduce *inference-friendly graph compression scheme* (IFGC), a as a general scheme to scale GNN inference to large graphs. We characterize IFGC with an *inference equivalence* relation, which captures the nodes with embeddings that are indistinguishable for the inference process using the same type of node update function. We then introduce a sufficient condition for the existence of IFGC, which specifies $G_c$ as the quotient graph of $G$ induced by the inference equivalence relation.

(2) We specify IFGC for representative GNNs classes. We first introduce structural preserving compression (SPGC), that enforces node embedding equivalence and neighborhood connectivity. We show it computes a compressed graph $G_c$ in $O(|E| \log |V|)$ time, which can be directly processed by the inference process to retrieve the original results *without* decompression. We further justify SPGC by showing that it can produce a unique, smallest $G_c$ up to graph isomorphism among compressed graphs. We also show that SPGC preserves the discriminability of the GNNs.

| Methods | Category | LB | MA | IFGS | Compression Cost |
|---------|----------|----|----|----|------------------|
| Dspar [37] | S | ✗ | ✓ | ✓ | $O(\frac{|V|log|V|}{\epsilon^2})$ |
| AdaptiveGCN [34] | S | ✓ | ✗ | ✗ | $n/a$ |
| NeuralSparse [56] | S | ✓ | ✓ | ✗ | $O(q|E|)$ |
| SCAL [29] | C | ✓ | ✓ | ✗ | $n/a$ |
| FGC [32] | C | ✓ | ✓ | ✗ | $O(|V|^2|V_c|)$ |
| SPGC (Ours) | C | ✗ | ✓ | ✓ | $O(|V| + |E|)$ |
| $(\alpha, r)$-SPGC (Ours) | C | ✗ | ✓ | ✓ | $O(|V||N_r| + |E|)$ |
| ASPGC (Ours) | C | ✗ | ✓ | ✓ | $O(|G_L|)$ |

**Table 1: Graph compression to accelerate GNN inference. S: Sparsification, C: Coarsening, LB: Learning-Based, MA: Model-Agnostic, IP: Inference-friendly w. guarantee, $\epsilon$: a constant that controls approximation error, $q$: # of visits of the neighbors per node. $N_r$: the largest $r$-hop neighbor set for a node in $G$. $G_L$: the subgraph of $G$ induced by $L$-hop of anchored node set $V_A$ for $L$-layered GNNs. .**

(3) We further introduce two *configurable* variants of IFGC to allow flexible trade-off between compression ratio and the quality of inference output. (a) The $(\alpha, r)$-SPGC groups nodes with similar features (determined by a threshold $\alpha$), that also have similar counterparts within their $r$-hop neighbors. (b) The *anchored*-SPGC (ASPGC) adapts SPGC to an "anchored" node set of user's interests and preserve the inference results for such nodes only rather than the entire node set. For both variants, we introduce efficient compression and inference algorithms.

(4) We experimentally verify the effectiveness and efficiency of our graph compression schemes. We show that with cheap "once-for-all" compression, our compression methods can significantly reduce the inference cost of representative GNNs such as GCNs, GAT and GraphSAGE by 55%-85%, with little to no sacrifice of their accuracy.

**Related work**. Several approaches have been developed to accelerate GNN inference in large graphs [14, 21, 41, 58]. Closer to our approach is graph reduction, which simplifies graphs at a small sacrifice of model accuracy [26, 35]. There are three strategies.

*Graph Sparsification*. These methods (learn to) remove task-irrelevant edges from input graphs, such that the remaining part preserves the performance of GNNs. For example, AdaptiveGCN [34] learns an edge predictor to determine and remove task-irrelevant edges to accelerate GNN inference on CPU/GPU clusters. NeuralSparse [56] learns supervised DNNs to remove task-irrelevant edges. [14] proposed a framework to incorporates both model optimization and graph sparsification, which leverages lottery ticket hypothesisto identify subnetworks that can perform as well as the full network. Dspar [37] induces smaller subgraphs by removing edges that have similar "importance" (quantified by approximating a resistance measure as in circuits) to preserve graph spectrum.

*Graph Coarsening*. These methods group and amalgamate nodes into groups, without removing nodes. For example, SCAL [29] proposed the use of off-the-shelf coarsening methods LV[39] for scaling up GNN training and theoretically proved that coarsening can be considered a type of regularization and may improve the generalization as well as reduce the number of nodes by up to a factor of ten without causing a noticeable downgrade in classification accuracy. [32] introduced an optimization-based framework (FGC) that incorporates graph matrix and node features to jointly learn a coarsened graph while preserving desired properties such as spectral similarity [39]. GRAPE [51] is a GNN variant enhanced with sampled subgraph features from ego networks of automorphic equivalent nodes. It has a different goal of improving accuracy rather than reducing inference costs. [10] compresses graphs to accelerate GNN learning, using color refinement (with a case of bisimulation) to merge nodes within bounded radius. Node groups are iteratively refined based on a label encoding that concatenate node label and neighboring group colors. This is similar with SPGC. Nevertheless, no inference algorithm is provided. We show that our $(\alpha, r)$-compression subsumes bisimulation compression.

*Graph Sketch*. These methods reduce the redundancy in the original graph, generating a skeleton graph that retains essential structural information. For instance, Graph-Skeleton [12] constructs a compact, synthetic, and highly-informative graph for the target nodes classification by eliminating redundant information in the background nodes. While this approach effectively reduces the memory usage for a pre-defined set of target nodes, it lacks flexibility and generalizability, as it must be tailored to specific nodes. In contrast, our SPGC performs one-time compression that applies universally to any set of test nodes $V_T \subset V$. NeutronSketch [36] focuses on eliminating the redundant information from the training portion of the graph, yet its compression is restricted to the training phase rather than accelerating inference, and does not ensure inference equivalence for the compressed graphs.

Our work differs from existing graph reduction approaches (summarized in Table. 1) in the following. (1) Our methods are model-agnostic and apply to any GNN that adopt the same inference process, without requiring model parameters, and incur no learning overhead. (2) We specify IFGC with variants that (approximately) preserve inferred results with invariant properties such as uniqueness and minimality, as well as fast compression and inference algorithms. These are not discussed in prior work. On the other hand, we remark that our scheme can be applied orthogonally: One can readily apply these approaches over compressed graphs from our method to further improve GNN training and inference. Our proposed SPGC can be potentially extended to the following learning settings: (1) transductive-learning: training on a compressed graph $G_c$ while utilizing our proposed memoization structure $\mathcal{T}$ to enable accuracy-preserving inference; and (2) inductive-learning: applying semi-supervised or unsupervised learning techniques to refine the graph structure of $G_c$ for a specific task.

## 2 GRAPHS AND GRAPH NEURAL NETWORKS

**Graphs**. A directed graph $G = (V, E)$ has a set of nodes $V$ and a set of edges $E \subseteq V \times V$. Each node $v$ carries a tuple $T(v)$ of attributes and their values. The size of $G$, denoted as $|G|$, refers to the total number of its nodes and edges, *i.e.*, $|G| = |V| + |E|$.

**Graph Neural Networks**. A graph neural network (GNN) $\mathcal{M}$ is a mapping that takes as input a featurized representation $G = (X, A)$ to an output embedding matrix $Z$, *i.e.*, $\mathcal{M}(G) = Z$. Here $X$ is a matrix of node features, and $A$ is a (normalized) adjacency matrix of $G$. [1]

---
[1]A feature vector $X_v$ of a node $v$ can be a word embedding or one-hot encoding [22] of $T(v)$. $A$ is often normalized as $\hat{A} = A + I$, where $I$ is the identity matrix.

| Notation | Description |
|---|---|
| $G = (X, A)$ | graph $G$, $X$: feature matrix, $A$: adjacency matrix |
| $|G|$ | size of $G$; $|G| = |V| + |E|$ |
| $\mathcal{M}$ | a GNNs model |
| $\mathcal{M}(G)$ (resp. $\mathcal{M}(G, V_T)$) | output of $\mathcal{M}$ over $G$ (resp. test set $V_T$) |
| $\mathcal{C}, \mathcal{P}$ | compression, post-processing function |
| $M_v$ | node update function |
| $x_v^k$ | embedding of node $v$ at layer $k$ |
| $G_c$ | compressed graph of $G$ |
| $\alpha, r$ | similarity threshold, # hops |
| $V_T, V_A$ | test node set, anchored nodes |
| $R^S, R^{(\alpha, r)}, R_c^A$ | structural equivalence, $(\alpha, r)$ relation & anchored relation |

**Table 2: Summary of Notations.**

**Inference.** We take a query language perspective [6, 23] to characterize the inference process of GNNs. A GNN inference process is specified as a composition of *node update functions*.

*Node update function.* Given a GNN $\mathcal{M}$ with $L$ layers, a node update function $M_v$ uniformly computes the embedding of each node $v$ at each layer $k$ ($k \in [1, L]$), with a general recursive formula as

$$x_v^k = M_v^k(\Theta^k, \text{AGG}(X_u^{k-1}, x_v^{k-1}, \forall u \in N^k(v)))$$

which is specified by (1) the learned model parameters $\Theta^k$, (2) an aggregation function AGG (*e.g.*, $\sum$, CONCAT), and (3) the neighbors of $v$ that participate in the inference computation at the $k$-th layer (denoted as $N^k(v)$). When $k=1$, $X_v^0 = X_v \in X$, *i.e.*, the input features.

The *inference process* of a GNN $\mathcal{M}$ with $L$ layers takes as input a graph $G = (X, A)$, and computes the embedding $x_v^k$ for each node $v \in V$ at each layer $k \in [1, L]$, by recursively applying the node update function. A GNN $\mathcal{M}$ has a *fixed* inference process, if its node update function is specified by fixed input model parameters, layer number, and aggregator. It has a *deterministic* inference process, if $M(\cdot)$ always generates the same embedding for the same input.

We consider GNNs with fixed, deterministic inference processes. In practice, such GNNs are desired for consistent and robust performance. For simplicity, we assume that $M_v$ specifies a proper set of neighbors that participate the inference process as $N(v) \subseteq \{u | (u, v) \in E$ or $(v, u) \in E\}$. This allows us to include GNNs that exploits neighborhood sampling (such as GraphSAGE), and directed message passing into discussion. In general, inferences of representative GNNs are in PTIME [13, 58] (see Table 3).

**Classes of** GNNs. We say a set of fixed, deterministic GNNs $\mathbb{M}$ belongs to a *class of* GNNs $\mathbb{M}^L$, if for every GNN $\mathcal{M} \in \mathbb{M}$, (1) $\mathcal{M}$ has $L$ layers, and (2) $\mathcal{M}$ uses the same form of node update function $M_v^k$, for each node $v \in V$ and $k \in [1, L]$.

Table 3 summarizes several node update functions in their general forms for mainstream GNN classes. For example, Graph Convolution Networks (GCNs) [31] adopt a node update function as $X_v^k = \sigma(\Theta^k(\sum_{u \in N(v)} \frac{1}{\sqrt{d_u d_v}} x_u^{k-1}))$. Here $d_u$ or $d_v$ denotes the degree of node $u$ or $v$. $\sigma(.)$ is the non-linear activation function. A class of GNNs $\text{GCN}^3$ contains 3-layered GCNs that adopt such node update function. Note that two GNNs in the same class can have different $\Theta$ and output, given the same input.

## 3 INFERENCE-FRIENDLY COMPRESSION

Given a graph $G = (V, E)$, a *compressed graph* of $G$, denoted as $G_c = (V_c, E_c)$, is a graph where (1) each node $[v] \in V_c$ is a nonempty subset of $V$, and $V = \bigcup_{[v] \in V_c} [v]$; and (2) there is an edge $([v], [v']) \in E_c$, if there are at least one node $v \in [v]$ and $v' \in [v']$, such that $(v, v') \in E$.

Note that $|V_c| \leq |V|$ and $|E_c| \leq |E|$. Hence, $|G_c| \leq |G|$.

**Inference-friendly Graph Compression.** Given a set of GNNs $\mathbb{M}$ and a graph $G$, an *inference-friendly graph compression*, denoted as IFGC, is a pair $(\mathcal{C}, \mathcal{P})$ where

- $\mathcal{C}$ is a compression function that computes a compressed graph $G_c$ of $G$ ($G_c = \mathcal{C}(G)$);
- $\mathcal{P}$ is a function that restore the auxiliary information of nodes in $G_c$ to their counterparts in $G$; and moreover,
- $\mathcal{M}(G) = \mathcal{M}(\mathcal{P}(G_c))$, for *any* GNN $\mathcal{M} \in \mathbb{M}$.

An IFGC aims to generate a compressed graph $G_c$ with a smaller size, such that an inference query that requests output $\mathcal{M}(G, V_T)$ for any $V_T \subseteq V$ can be computed by a faster inference process of $\mathcal{M}$ over $G_c$ only, even with a query-time overhead incurred by $\mathcal{P}$.

**A Sufficient Condition.** We next introduce a sufficient condition for the existence of IFGC. To this end, we start with a notion of *inference-equivalent* relation.

**Inference equivalence.** Given a class of GNN $\mathbb{M}^L$ and a graph $G$, a pair of nodes $(v, v')$ in $G$ are *inference equivalent w.r.t.* $\mathbb{M}^L$, denoted as $v \sim_\mathcal{M}^L v'$, if for any $\mathcal{M} \in \mathbb{M}^L$, $X_v^k = X_{v'}^k$ for any $k \in [1, L]$.

One can readily infer that for any two nodes $v \sim_\mathcal{M}^L v'$, $M(v, G) = M(v', G)$. That is, inference equivalence of nodes ensure that they are all "indistinguishable" for the inference of any GNN $\mathcal{M} \in \mathbb{M}^L$.

Denote the binary relation $(v, v')$ induced by inference equivalence as $R_\mathcal{M}^L$, *i.e.*, $(v, v') \in R_\mathcal{M}^L$ if and only if $v \sim_\mathcal{M}^L$. We say $R_\mathcal{M}^L$ is *nontrivial* if there is at least one pair $(v, v') \in R_\mathcal{M}^L$, where $v \neq v'$. We can readily verify the following result.

**Lemma 1:** *Given $\mathbb{M}$ and $G$, the binary relation $R_\mathcal{M}^L$ is an equivalence relation*, i.e., *it is reflexive, symmetric, and transitive.* □

The *equivalent class* of $v$ under an equivalence relation $R_\mathcal{M}^L$, denoted as $[v]$, refers to the set $\{v' | (v, v') \in R_\mathcal{M}^L\}$. The equivalent classes induced by the inference equivalence relation $R_\mathcal{M}^L$ forms a node partition $V_R$ of $V$. The *quotient graph* induced by $R_\mathcal{M}^L$ is a graph $G_R$ with nodes $V_R$ and edges $E_R$, where each node in $V_R$ is a distinct equivalent class induced by $R_\mathcal{M}^L$, and there is an edge $([v], [v']) \in E_R$ if and only if there exists a node $v \in [v]$ and $v' \in [v']$, such that $(v, v') \in E$.

**Lemma 2:** *Given a class of GNNs $\mathbb{M}^L$ and a graph $G$, a graph compression scheme $(\mathcal{C}, \mathcal{P})$ is an IFGC w.r.t. $\mathbb{M}^L$ and $G$, if for any $\mathcal{M} \in \mathbb{M}^L$, (1) $\mathcal{C}(G)$ computes a quotient graph $G_c$ induced by a non-trivial inference equivalent relation $R_\mathcal{M}^L$ w.r.t. $\mathbb{M}^L$ and $G$, and (2) $\mathcal{P}$ is a function that restores $X_v^k$ with $X_{[v]}^k$ by a scaling factor derived from auxiliary information of $v$, for each layer $k \in [1, L]$.* □

**Proof sketch:** Let $R_\mathcal{M}^L$ be a non-empty inference equivalent relation *w.r.t.* $\mathbb{M}^L$ and $G$, and $G_c$ be the quotient graph induced by $\mathbb{M}^L$. (1) Given **Lemma 1**, $R_\mathcal{M}^L$ is a nontrivial equivalence relation. Hence there exists at least one equivalent class $[v]$ with size larger than one, i.e., $|G_c| < |G|$. As function $\mathcal{P}$ does not introduce new node or edge to $G_c$, we have $|C(G))| = |G_c| < |\mathcal{P}(G_c)| < |G|$. (2) To see $\mathcal{M}(G) = \mathcal{M}(\mathcal{P}(C(G)))$, i.e., $G_c$ preserves inference result, it suffices to show that for every node $v \in G$, $\mathcal{M}(G_c, \{v\}) = M(\mathcal{P}(C(G)), \{[v]\})$. This is ensured by (a) the fixed deterministic inference process that applies the same node update function $M_v$, and (b) $\mathcal{P}$ restores $X_v^k$ with only $X_{[v]}^k$ and a scaling factor, for any layer $k \in [1, L]$. We list examples of $\mathcal{P}$ and scaling factors for mainstream GNNs in Table 4. Hence $(C, \mathcal{P})$ is an IFGC. □

| GNNs Classes | Node Update Function (general form) | Training Cost | Inference Cost |
|---|---|---|---|
| Vanilla [45] | $X_v^k = \sigma(\Theta \cdot \text{AGG}(X_u^{k-1}, \forall u \in \mathcal{N}(v)))$ | $O(L|E| + L|V|)$ | $O(L|E| + L|V|)$ |
| GCN [13, 31] | $X_v^k = \sigma(\Theta^k(\sum_{u \in \mathcal{N}(v)} \frac{1}{\sqrt{d_u d_v}} x_u^{k-1}))$ | $O(L|E|F + L|V|F^2)$ | $O(L|E|F + L|V|F^2)$ |
| GAT [9, 49] | $X_v^k = \sigma(\sum_{u \in \mathcal{N}(v)} \alpha_{uv} \Theta^k X_v^{k-1})$ | $O(L|E|dF^2 + L|V|F^2)$ | $O(L|E|dF^2 + L|V|F^2)$ |
| GraphSAGE [13, 24] | $X_v^k = \sigma(\Theta^k \cdot (X_v^{k-1} \| \text{AGG}(X_u^{k-1}, \forall u \in \mathcal{N}(v))))$ | $O(L|V|dF + L|V|F^2)$ | $O(L|V|dF + L|V|F^2)$ |
| GIN [9, 52] | $X_v^k = \sigma(\text{MLP}((1 + \gamma)x_v^{k-1} + \sum_{u \in \mathcal{N}(v)} x_u^{k-1})$ | $O(L|E|F + L|V|F^2)$ | $O(L|E|F + L|V|F^2)$ |

**Table 3: Comparison of Representative GNNs** with node update functions, training cost, and inference cost. $\sigma$: an activation function e.g., ReLU or LeakyReLU. AGG: aggregation function; can be e.g., sum ($\sum$), average (Avg), or concatenation ($\|$). $L$, $|E|$, $|V|$, $F$, and $d$ denote the number of layers, edges, nodes, features per node, and maximum node degree of $G$, respectively.

We next introduce practical IFGC for representative GNN classes, with efficient compression (implementing $C$) and inference (involving $\mathcal{P}$) algorithms. We summarize notations in Table 2.

## 4 STRUCTURAL-PRESERVING COMPRESSION

We introduce a first IFGC for GNN inference. We specify $R_M^L$ as an extended version of *structural equivalence*. The latter has origins in role equivalence in social science [38], and simulation equivalence of Kripke structures in model checking [4, 16]. By enforcing equivalence on embeddings and neighborhood connectivity, it ensures an IFGC to accelerate GNN inference *without* decomposition.

### 4.1 Compression Scheme

**Structural equivalence.** Given a graph $G = (X, A)$, a *structural equivalence* relation, denoted as $R^S$, is a non-empty binary relation such that for any node pair $(v, v')$ in $G$, $(v, v') \in R^S$, if and only if:

○ $X_v^0 = X_{v'}^0$, i.e., $v$ and $v'$ have the same input features;

○ for any neighbor $u$ of $v$ ($u \in \mathcal{N}(v)$), there exists a neighbor $u'$ of $v'$ ($u'$ in $\mathcal{N}(v')$), such that $(u, u') \in R^S$; and

○ for any neighbor $u''$ of $v'$ in $\mathcal{N}(v')$, there exists a neighbor $u'''$ of $v$ in $\mathcal{N}(v)$, such that $(u'', u''') \in R^S$.

**Structural-preserving Compression.** Given a GNN class $\mathbb{M}^L$ and graph $G$, a *structural-preserving compression*, denoted as SPGC w.r.t. $\mathbb{M}^L$, is a pair $(C, \mathcal{P})$ where (1) $C$ computes $G_c$ as the quotient graph of $R^S$, where $R^S$ is the non-empty, maximum structural equivalence relation in $G$, and (2) $\mathcal{P}$ is a function that restores node embeddings with matching scaling factors for $\mathbb{M}^L$.

**Example 2:** Consider the graphs $G_2$ and $G_3$ in Fig. 2, and their compressed counterpart obtained by SPGC, $G_{c_2}$ and $G_{c_3}$, respectively.

(1) $G_2$ has 10 nodes and 10 edges. The nodes having the same labels $'a', 'c', 'd'$ also have the same input features, respectively. For example, $X_{a_1}^0 = X_{a_2}^0$, and $X_{d_1}^0 = X_{d_2}^0 = X_{d_3}^0$. For nodes labeled with $'b'$, $X_{b_1}^0 = X_{b_3}^0 \neq X_{b_2}^0$. One can verify that $R^S = \{(b_1, b_3), (c_1, c_2), (d_1, d_3)\}$. A compressed graph $G_{c_2}$ is illustrated with 7 nodes and 7 edges.

Observe that despite $d_1$, $d_2$ and $d_3$ have the same input features, $d_2 \not\sim_M^L d_1$, and $d_2 \not\sim_M^L d_3$ for GNNs with $L \geq 1$. Indeed, $d_2$ has a neighbor $b_2$ that has no counterpart in the neighbors of $d_1$ or $d_3$ that share the same embedding; hence the output embedding of $d_2$ may be different from either $d_1$ or $d_3$, and should be separated from equivalent class $\{d_1, d_3\}$ in $G_c$.

(2) $G_3$ is a cycle in the form of $\{c_n, b_n, a_n, \ldots, c_1, b_1, a_1\}$, where $X_{a_i}^0 = X_{a_j}^0$, $X_{b_i}^0 = X_{b_j}^0$, and $X_{c_i}^0 = X_{c_j}^0$, for any $i, j \in [1, n]$. We can verify that $R^S = \bigcup_{i,j \in [1,n]} \{(a_i, a_j), (b_i, b_j), (c_i, c_j)\}$. A smallest compressed graph $G_{c_3}$ is illustrated with only three nodes: $A = \{a_i\}$, $B = \{b_i\}$, $C = \{c_i\}$, $\forall i \in [1, n]$, regardless of how large $n$ is. □

The result below tells us that any two nodes that are structural equivalent are "indistinguishable" for GNN inference process.

**Theorem 3:** *Given a class of GNNs $\mathbb{M}^L$ and graph $G$, the relation $R^S$ over $G$ is an inference equivalence relation* w.r.t. $\mathbb{M}^L$. □

**Proof sketch:** First, $R^S$ is an equivalence relation. It then suffices to show that for any pair $(v, v') \in R^S$, $v \sim_M^L v'$. We perform an induction on the number of layers $k$ for GNNs. Consider a "matching" relation $h$ between a pair $(v, v') \in R^S$, such that $h(v) = v'$. At any layer, for any node $v$ and every neighbor $u \in \mathcal{N}(v)$, there exists a "match" $h(v)$ and a "match" $h(u) \in \mathcal{N}(h(v))$ with the same (intermediate) embedding. This ensures the equivalence of aggregated embedding computed by the node update function at $v$ and $h(v)$, and vice versa. Hence for any pair $(v, v') \in R^S$, $v \sim_M^L v'$. $R^S$ is thus an inference-friendly relation by definition. □

Following Lemma 2 and Theorem 3, SPGC is an IFGC.

**Example 3:** We also compare SPGC with several other possible graph compression scheme. We illustrate three more compressed graphs, $G_{c_2}^1$, $G_{c_2}^2$, $G_{c_2}^3$ of $G_2$, following Exact Compression [10], Bisimulation [16], and Automorphism [51], respectively.

(1) Exact Compression [10] applies an iterative color refinement process[2] starting with groups that contains nodes agreeing on embeddings and color encoding (labels). It then iteratively split the groups, where each node is updated by concatenating its color encoding with those from their neighbors, and refine groups. It finally derives $G_{c_2}^1$ with 8 nodes and 9 edges after two rounds, where only node pairs $d_1, d_3$ and $a_1, a_2$ share the same node representation. The concatenation is more sensitive to the impact of e.g., degrees, preventing more possible merge, hence less can be compressed.

(2) Bisimulation [16] ignores embedding equivalence and groups nodes only with topology-level equivalence. In this case, $b_2$ can be merged with $b_1$ and $b_3$ due to bisimulation in connectivity. This leads to smaller compressed graph $G_{c_2}^2$ with 4 nodes and 3 edges compared with $G_{c_2}$, yet at the cost of inaccurate inference at *e.g.*, nodes $c_1$, $c_2$, $b_2$, and $d_2$ due to "overly" compressed structure.

---

[2]The original method is used to simplify GNNs learning problems [10]; we make a comparison by applying color refinement for graph compression alone.

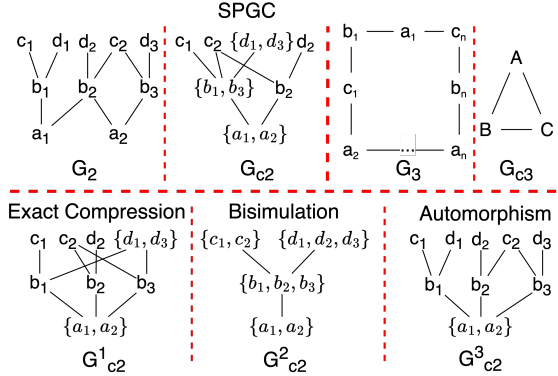

**Figure 2: Compressing graphs with** SPGC **and a comparison with exact compression [10], bisimulation [16], and automorphism (adapted from [51]). Exact compression uses node features for initial coloring and** $d = 2$ **for color refinement.**

(3) Automorphism [51] partitions nodes into same automorphism equivalence sets, which poses strong topological equivalence on graph isomorphism in their neighbors. By enumerating the automorphism groups of $G_2$ and considering the embedding similarities, only $a_1$ and $a_2$ can be merged as shown in $G_{c2}^3$ while other node pairs like $c_1, d_1$ or $b_1, b_3$ cannot be merged due to different embeddings or different connection patterns of the neighbors. This can be an overkill for reducing unnecessary inference computation. In addition, computing automorphism remains NP-hard, which indicates more expensive compression cost; while SPGC computes maximum structural equivalence in PTIME (see Section 4.3). □

### 4.2 Properties and Guarantees

We next justify SPGC by showing a *minimality and uniqueness* property. We show that an SPGC can generate a smallest $G_c$, which is "unique" up to graph isomorphism. That is, if there is another smallest compressed graph $G'_c$ generated by an SPGC, then $G_c^*$ and $G'_c$ are isomorphic.

**Lemma 4:** *Given a GNN class $\mathbb{M}^L$ and $G$, there is an SPGC that computes a smallest $G_c^*$ which is unique up to graph isomorphism.* □

**Proof sketch:** We show the minimality property by a constructive proof as follows: (1) given $\mathbb{M}^L$ and $G$, there exists a unique, largest inference equivalence relation $R^{*S}$; (2) we construct an SPGC that computes $G_c^*$ as the quotient graph of $R^{*S}$. The uniqueness of $G_c^*$ can be shown by a contradiction: if there exists another smallest $G'_c$ that is not graph isomorphic to $G_c^*$, then either $G'_c$ is not smallest in sizes, or $R^{*S}$ is not the (largest) inference equivalence relation, *i.e.*, there is a pair $(v, v')$, such that either $v \sim_M^R v'$ but are not in $[v]$, or $v \nsim_M^R v'$, but are included in $[v]$. Either leads to contradiction. □

We next justify SPGC by showing that it properly preserves the discriminative set of GNNs, which has been used as one way to characterize the expressiveness power of GNNs as queries [6, 23].

**Discriminative set of** GNNs [23]. Given a set of graphs $\mathcal{G}$, the *discriminative set* of a GNN $\mathcal{M}$, denoted as $\mathcal{G}_\mathcal{M}$, refers to the maximum set of pairs $\{(G, G')\}$, where $G, G' \in \mathcal{G}$, such that $M(G) = M(G')$. In the case of equivariant GNNs [44], the strongest discriminativeness can be achieved, for which the set contains all pairs $(G, G')$ such that $G$ and $G'$ are isomorphic [5]. In other words, these GNNs can "solve" graph isomorphic problem: one can issue a Boolean inference query to test if an input pair of graphs is isomorphic.

Given a set of graphs $\mathcal{G}$, denote the set of corresponding compressed graphs generated by SPGC as $\mathcal{G}_c$, *i.e.*, $\mathcal{G}_c = \{G_c | G_c = C(G); G \in \mathcal{G}\}$. We have the following result.

**Lemma 5:** *Given $\mathbb{M}^L$ and a set of graphs $\mathcal{G}$, an SPGC can compute a compressed set $\mathcal{G}_c$, such that for every GNN $M \in \mathbb{M}^L$, and any pair $(G, G') \in \mathcal{G}_\mathcal{M}$, there exists a pair $(G_c, G'_c) \in \mathcal{G}_{c_\mathcal{M}}$.* □

This result tells us that SPGC "preserves" the discriminativeness of GNNs. Moreover, it suggests a practical compression scheme for large-scale graph classification. One can apply SPGC to compress $\mathcal{G}$ to a smaller counterpart $\mathcal{G}_c$. As the discriminativeness set is preserved over $\mathcal{G}_c$ for every GNN $M \in \mathcal{M}^L$, SPGC reduces the overall classification overhead, via a post-processing $\mathcal{P}$ that readily groups $\mathcal{G}$ by corresponding label groups over $\mathcal{G}_c$.

Due to limited space, we present the detailed proofs in [1].

### 4.3 Compression Algorithm

We next present a compression algorithm (function $C$) in SPGC.

**General idea.** The algorithm, simply denoted as SPGC, follows Lemma 4 to construct the smallest $G_c^*$ induced by the maximum structure equivalence relation $R_S^*$. To ensure efficient inferences that only refer to $G_c$ without decompression, it (1) uses a *memoization structure* $\mathcal{T}$ to cache the neighborhood statistics specified by node update function $M_v$, and (2) *rewrites* $M_v$ to an equivalent counterpart $M_{[v]}$ (see Table 4 for examples), such that the inference can directly process on each $[v]$ in $G_c$, and "looks up" $\mathcal{T}$ at runtime, to obtain the embeddings for all the nodes in $[v]$, in a single batch.

**Compression Algorithm.** The SPGC algorithm, as illustrated in Fig. 3, takes as input a featurized input $G = (X, A)$ and a GNN class $\mathbb{M}^L$ with node update function $M_v$. (1) It first extends Dovier-Piazza-Policriti (DPP) algorithm [16] to compute the maximum structural equivalence relation $R_S^*$, by enforcing embedding equivalence as an additional equivalence constraint (lines 2-4). This induces a set of equivalence classes $EC$ (a node partition). It then invokes a procedure CompressG to construct $G_c$ as the quotient graph of $R_S^*$, as well as the memoization structure $\mathcal{T}$ (line 5).

**Procedure** CompressG. CompressG is a light-weighted compression approach enabling an easy-to-implement and feasible way

---

**Algorithm 1 :** SPGC

**Input:** Graph $G$, node feature matrix $X$, a class of GNNs $\mathbb{M}^L$ with node update function $M_v$;

**Output:** A compressed graph $G_c$ with memoization structure $\mathcal{T}$;

1: set $R^S := \emptyset$; set $EC := \{V\}$; set $\mathcal{T} := \emptyset$; graph $G_c := \emptyset$;
2:    $R^S := \text{DPP}(G)$;
3:    $R^S := R^S \setminus \{(v, v') \mid X_v^0 \neq X_{v'}^0\}$;
4:    $EC := V/R^S$; /* induce partition $EC$ from refined $R^{S*}$*/
5:    $(G_c, \mathcal{T}) := \text{CompressG }(\mathcal{T}, G_c, EC, G, M)$;
6:    **return** $G_c$ and $\mathcal{T}$;

**Figure 3: Algorithm** SPGC

---

**Algorithm 2** Procedure CompressG($\mathcal{T}$, $G_c$, $EC$, $G$, $M$)

---

1: **for** $[v] \in EC$ **do**
2:      $V_c = V_c \cup \{[v]\}$;
3:      initialize $[v]_\mathcal{T}$; /* with row pointers as $v \in [v]$*/
4: **for** edge $(u, v) \in E$ **do**
5:      $E_c = E_c \cup \{([u], [v])\} \mid u \in [u], v \in [v]$;
6:      **if** $M.\phi$ is topology sensitive **then**
7:          $[v]_\mathcal{T}(v, [u]) \mathrel{+}= \frac{1}{\sqrt{deg(u)}}$;
8:      **else if** $M.\phi$ is weight sensitive **then**
9:          $[v]_\mathcal{T}(v, [u]) \mathrel{+}= \alpha_{v,u}$;
10:      **else**
11:          $[v]_\mathcal{T}(v, [u]) \mathrel{+}= 1$;
12:      $\mathcal{T} = \bigcup_{[v] \in V_c} [v]_\mathcal{T}$;
13: **return** $G_c$ and $\mathcal{T}$;

---

**Figure 4: Procedure** CompressG

to derive compressed graph $G_c$ from large-scale $G$ while computing the memoization structure $\mathcal{T}$. Given the induced equivalence classes $EC$ and an encoding of the node update function $M_v$, procedure CompressG (illustrated in Fig. 4), generates the compressed graph $G_c$ and memoization structure $\mathcal{T}$. For each equivalent class $[v]$ in $EC$, CompressG initializes a node in $G_c$, (lines 1-3). For each edge $(u, v) \in E$, CompressG adds an edge ($[u]$, $[v]$) (lines 4-5).

*Compression time memoization.* CompressG dynamically maintains a memoization structure $\mathcal{T}$ that is shared by all GNNs in $\mathcal{M}^L$, to cache useful auxiliary neighborhood information used by the node update function for efficient inference (see Section 4.4). For each node $[v] \in V_c$, it assigns $[v]_\mathcal{T}$, a compact table, such that for every $v \in [v]$, and every neighbor $[u] \in N([v])$, an entry $[v]_\mathcal{T}(v, [u])$ records an aggregation of auxiliary neighborhood information (*e.g.*, sum of node degree, edge weights) of $N(v) \subseteq N([v])$.

When processing an edge $(u, v) \in E$, it follows a case analysis of $\mathbb{M}^L$ with node update function $M_v$. For example, (1) "topology sensitive" means the degrees of neighbors of $v$ is required, as seen in GCNs; (2) "weight sensitive" means additional edge weights, such as edge attentions in GATs. For GATs, the edge weights from pre-trained $M$ are optional such that if given the information of all edge weights, the inference accuracy can be preserved based on the formula shown in Table. 4.4. Such information can be readily obtained by tagging the input GNN class $\mathbb{M}^L$ or encoded as rules. It then updates the entry $[v]_\mathcal{T}(v, [u])$ accordingly (lines 6-12).

**Example 4:** Consider a GNNs class GIN, and graph $G_4$ shown in Fig. 5. (1) SPGC invokes the DPP algorithm to compute $R^S$. It next refines $R^S$ based on feature embeddings and returns the induced partition $EC = \{[a], [b], [c], [d]\}$, where nodes with same labels are merged, *e.g.*, $[a] = \{a_1, a_2\}$. (2) We illustrate how CompressG dynamically updates the memoization structure $\mathcal{T}$ by considering the processing of two edges $(b_1, a_1)$ and $(b_2, a_1)$. For $[a] \in EC$, it first initializes $[a]_\mathcal{T}$ as an empty table. It next iterates over edges in $E$. As the node update function of GIN does not require exact degree or additional edge weight (not topology or weight sensitive), the entry $[a]_\mathcal{T}(a_1, [b])$ is updated to 1, to "memoize" that there is one neighbor of $a_1$ in $N([b])$ that will contribute to a "unit" value to the embedding computation of $a_1$, via edge $(b_1, a_1)$. Similarly, when it reaches edge $(b_2, a_1)$, $[a]_\mathcal{T}(a_1, [b])$ is updated to 2. Following this processes, all entries in $\mathcal{T}$ will be updated to memoize neighbors' information while compressing the graph. □

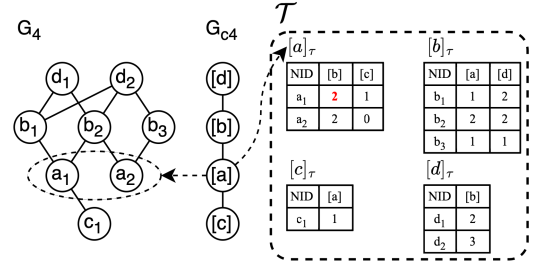

**Figure 5: Run-time generation of Memoization structure $\mathcal{T}$.**

**Correctness and cost.** SPGC correctly computes $G_c^*$ as ensured by (1) the correctness of DPP algorithm and (2) the follow-up refinement by enforcing embedding equivalence. For time cost, it takes SPGC $O(|V| + |E|)$ time to initialize $R^S$ with DPP algorithm. The refinement of $R^S$ and $EC$ takes $O(|V|)$ time (lines 3-4). Procedure CompressG processes each equivalent class in $EC$ ($|EC| \le |V|$) and each edge in $G$ once, hence in $O(|V| + |E|)$ time to construct $G_c^*$ and update $\mathcal{T}$. The total cost is thus in $O(|V| + |E|)$ time.

### 4.4 Inference Process

**Inference algorithm.** We outline an algorithm that directly obtains $M(G)$ by referring to $G_c^*$ only, without decompression. Our strategy rewrites the node update function $M_v$ to an equivalent counterpart $M_{[v]}$, that takes as input $[v]$ and the corresponding tuple $[v]_\mathcal{T}(v)$ in $\mathcal{T}$, to "scale" the embedding computation with the memorized edge weights. The algorithm performs inference directly in $G_c^*$ with $M_{[v]}$, and simply "scale up" the results at $[v]$ for each node $v \in [v]$, with a scaling factor. The scaling factor can be directly looked up from the table $[v]_\mathcal{T}(v)$ (function $\mathcal{P}$). Table 4 illustrates the scaling factors for mainstream GNN classes.

**Example 5:** Continuing with Example. 4, an inference at node $a_1$ looks up, in constant time, the values from entries $[a]_\mathcal{T}(a_1, [b])$ and $[a]_\mathcal{T}(a_1, [c])$ which are 2 and 1 separately (as shown in Fig. 5). It next assigns the values as scaling factors (Table. 4) to restore messages and the embedding of node $a_1$ as in original $G_4$. □

**Inference cost.** As SPGC requires no decompression on neighborhood structures of nodes, an inference query can be directly applied to $G_c$ without incurring additional overhead. The overall inference cost is in $O(L|E_c|F + L|V_c|F^2)$. We remark that this result is derived by scaling down a common upper bound of inference costs for mainstream GNNs in Table 3. For other and more complex GNNs variants, the inference costs can be derived similarly by scaling down from their counterparts over $G$.

## 5 CONFIGURABLE GRAPH COMPRESSION

While SPGC generates $G_c$ that can be directly queried by inference queries without decompression, it enforces node embedding equivalence, which may be an overkill for nodes with similar embeddings and can be processed in a single batch with tolerable difference in query outputs. Users may also want to *configure* the compression schemes to balance among accuracy and speed up, or to contextualize the compression with inference queries that specifies a set of test nodes $V_T \subseteq V$ of interests, such that $M(G, V_T) = M(G_c, V_T)$.

| GNNs | Node Update Function $M_v$ | equivalent rewriting $M_{[v]}$; scaling factors are marked in red | notes |
|---|---|---|---|
| Vanilla [45] | $X_v^k = \sigma(\Theta \cdot \text{AGG}(X_u^{k-1}, \forall u \in \mathcal{N}(v)))$ | $X_v^k = \sigma(\Theta \cdot \text{AGG}(\textcolor{red}{[\mathbf{v}]_\mathbf{T}(v, [\mathbf{u}])}X_{[u]}^{k-1}, \forall [u] \in \mathcal{N}([v])))$ | AGG: $\sum$ or AVG; for $AVG$, need to multiply by $RF_v$ |
| GCN [31] | $X_v^k = \sigma(\Theta^k(\sum_{u \in \mathcal{N}(v)} \frac{1}{\sqrt{deg_u deg_v}} x_u^{k-1}))$ | $X_v^k = \sigma(\Theta^k(\sum_{[u] \in \mathcal{N}([v])} \textcolor{red}{\frac{1}{\sqrt{deg_v}}[\mathbf{v}]_\mathbf{T}(v, [\mathbf{u}])}x_{[u]}^{k-1}))$ | $deg_v$: degree of node $v$ in $G$, topology sensitive |
| GAT [49] | $X_v^k = \sigma(\sum_{u \in \mathcal{N}(v)} a_{uv}\Theta^k X_u^{k-1})$ | $X_v^k = \sigma(\sum_{[u] \in \mathcal{N}([v])} \textcolor{red}{[\mathbf{v}]_\mathbf{T}(v, [\mathbf{u}])}\Theta^k X_{[u]}^{k-1})$ | weight sensitive |
| GraphSAGE [24] | $X_v^k = \sigma(\Theta^{k-1} \cdot |\text{AGG}(X_u^{k-1}, \forall u \in \mathcal{N}(v))))$ | $X_v^k = \sigma(\Theta^k \cdot (X_v^{k-1}||\text{AGG}(\textcolor{red}{RF_v \times [\mathbf{v}]_\mathbf{T}(v, [\mathbf{u}])}X_{[u]}^{k-1},$ $\forall [u] \in \mathcal{N}([v]))))$ | $||$: concatenation; AGG: AVG |
| GIN [52] | $X_v^k = \sigma(\text{MLP}((1+\gamma)x_v^{k-1} + \sum_{u \in \mathcal{N}(v)} x_u^{k-1}))$ | $X_v^k = \sigma(\text{MLP}((1+\gamma)x_{[v]}^{k-1} + \sum_{[u] \in \mathcal{N}([v])} \textcolor{red}{[\mathbf{v}]_\mathbf{T}(v, [\mathbf{u}])}x_{[u]}^{k-1}))$ | |

**Table 4: Rewriting of node update functions for mainstream GNN classes (scaling factors highlighted in <span style="color:red">red</span>).**

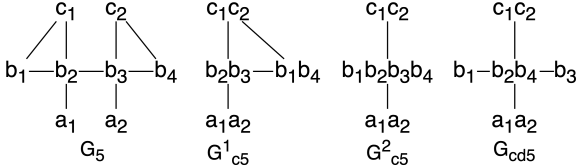

**Figure 6: Compressing $G_5$ with $(0.5, 2)$-SPGC.**

In response, we next introduce two variants of SPGC: $(\alpha, r)$-SPGC (Section 5.1), and anchored SPGC (Section 5.2), respectively.

## 5.1 Compression with Structural and Embedding Similarity

We start with a relation called $(\alpha, r)$-relation, which approximates $R^S$ by lifting its equivalence constraints.

$(\alpha, r)$**-relation.** Given graph $G$, a configuration $(\text{xsim}, \alpha, r)$ is a triple where $\text{xsim}(\cdot)$ is a *feature similarity function* that computes a similarity score of two node embeddings, $\alpha$ is a similarity threshold ($\alpha \in [0, 1]$), and $r$ an integer. Let $N_r(v)$ be the nodes within $r$-hop neighbors of $v$. A binary relation $R^{(\alpha, r)} \subseteq V \times V$ is an $(\alpha, r)$-relation if for any node pair $(v, v') \in R^{(\alpha, r)}$,
- $\text{xsim}(X_v^0, X_{v'}^0) \geq \alpha$;
- for any node $u \in N(v)$, there exists a node $u' \in N_r(v')$, such that $(u, u') \in R^{(\alpha, r)}$; and
- for any node $u'' \in N(v')$, there exists a node $u''' \in N_r(v)$, such that $(u'', u''') \in R^{(\alpha, r)}$.

Note that $R^{(1,1)}$ is an $R^S$, as $\alpha = 1$ ensures embedding equivalence, and $r = 1$ preserves indistinguishable neighbors for node update functions in GNN inference. On the other hand, $(\alpha, r)$-relation is no longer an equivalence relation, as it "relaxes" structural equivalence by lifting both embedding equality, and the strict neighborhood-wise equivalence, in trade for better compression ratio. We next clarified the relationship between $r$ and $L$. Based on the definition of SPGC, the embedding similarity between two vertices is dependent on $r$ instead of $L$. In other words, SPGC is model-agnostic as long as $r \leq L$. In practice, since $L$ is small due to over-smoothing issue of larger $L$, $r$ is usually a small integer like 1 or 2.

Based on the relation $R^{(\alpha, r)}$, we introduce a variant of SPGC.

$(\alpha, r)$-SPGC. Given a graph $G$, and a configuration $\alpha$ and $r$ w.r.t. an embedding similarity measure and a threshold, an $(\alpha, r)$-SPGC is a graph compression scheme if $C$ computes a graph $G_c$ induced by the relation $R^{(\alpha, r)}$. Specifically,
- for any node pair $(v, v') \in R^{(\alpha, r)}$, $v \in [v]$, $v' \in [v]$; and
- there is an edge between $([u], [v])$ if $(u, v) \in E$.

**Lemma 6:** *Given a GNN class $\mathcal{M}^L$ and a graph $G$, an $(\alpha, r)$-SPGC incurs compression cost in $O(|V||N_r| + |E|)$ time ($|N_r|$ refers to the largest size of $r$-hop neighbors of a node in $G$), and an inference cost in $O(L|E|F + L|V_c|F^2)$ time.* □

As a constructive proof, we next introduce algorithms that implements $(\alpha, r)$-SPGC with the above guarantees.

**Compression Algorithm.** We describe the compression algorithm $(\alpha, r)$-SPGC. It follows the same principle to compute $(\alpha, r)$-relation and induce a compressed graph. The difference is that rather than inducing equivalence class and quotient graph from $G$, (1) it first induces a graph $G_r$ by linking nodes to their $r$-hop neighbors, (2) it then computes an $R^S$ relation by invoking DPP algorithm, and refines it by a re-grouping of nodes determined by similarity function $\text{xsim}(\cdot)$ with $\alpha$ as similarity threshold. (3) It generates $[v]$ to include all the pairs $(v, v')$ in $R^{(\alpha, r)}$, and accordingly the edges. It updates the memoization structure $\mathcal{T}$ following the edges in $G_r$, similarly as in SPGC. Here $\mathcal{T}$ caches the statistics from the $r$-hop neighbors of each node $v$ in the original graph $G$.

**Example 7:** Consider graph $G_5$ shown in Fig. 6. A $(0.5, 2)$-SPGC invokes DPP to initialize a $(1, 1)$-relation, and refines it to $R^{(0.5,2)} = \{(a_1, a_2), (c_1, c_2), (b_1, b_2), (b_1, b_3), (b_1, b_4), (b_2, b_3), (b_2, b_4), (b_3, b_4)\}$. This yields a compressed graph $G_{c_5}^2$ with only 3 nodes and 2 edges.

We also illustrate $G_{c_5}^1$, a compression graph from SPGC for $G_5$ (induced by an $R^S$ as a $(1, 1)$-relation). Due to strictly enforced embedding equivalence, $b_1$ and $b_2$ cannot be merged, and similarly for $b_3$ and $b_4$. This yields $G_{c_5}^1$ with more nodes and edges. □

*Compression Cost.* It takes $O(|V| \cdot |N_r(v)|)$ time to derive $G_r$ for $v \in V$. It then takes $O(|V| + |E_r|)$ to compute and refine $R^{(\alpha, r)}$, for $G_r$ with edge set $E_r$. Here $|E_r| \leq |V| \cdot |N_r|$, where $N_r$ refers to the largest $r$-hop neighbor set for a node in $G$. Procedure CompressG constructs $G_c$ in $O(|V| + |E_r|)$ time, and generates memoization structure $\mathcal{T}$ in $O(|E|)$ time. The total cost is thus in $O(|V||N_r| + |E|)$.

As $(\alpha, r)$-SPGC is specified by $R^{(\alpha, r)}$ that approximates an inference-friendly relation, it is no longer an IFGC, hence a direct inference over the $G_c$ from it may not preserve the original output. To mitigate accuracy loss, the inference specifies a procedure $\mathcal{P}$ to perform run-time decompression with small overhead.

**Inference process with decompression**. The inference algorithm directly processes each node $[v]$ in $G_r$ as in SPGC. For $(\alpha, r)$-SPGC. the difference is that it ad-hocly invokes a decompression procedure decompG (the decompG algorithm and its example illustrated in Figure. 18 are shown in the Appendix [1]) to reconstruct the neighbors of $v$ in $G_r$, and performs an inference using original 1-hop neighbors of $v$ to obtain an embedding $X_v$ as close as its original counterpart in $G_r$. To minimize decompression cost, when compressing the graph, algorithm $(\alpha, r)$-SPGC (shown in Appendix B) incorporates *Re-Pair*, a reference encoding method [15, 33], to derive $AL_c$ and rules for later fast decompression from a compact encoding structure. Specifically, within each $EC$, procedure decompG sorts the processing order of nodes by their degrees in $G_r$. In other words, procedure decompG prioritizes the decompression of nodes having the most shared $r$-hop neighbors with others in $G_r$, to (1) maximally reduce redundant computation in decompression process, and (2)"maximize" the likelihood for more accurate inference computation. For example, a partially decompressed graph $G_{cd_5}$ that resolves 1-hop neighbors of $b_2$ (in trade for more accurate embedding) is illustrated in Fig. 6 (please refer to Example. 8 and Figure. 18 in Appendix for later [1]). The decompressed neighbors are kept in $G_{cd}$ until the inference terminates.

As the decompression restores at most $|E|$ edges, the overall inference process takes $O(L|E|F + L|V_c|F^2)$ time, including the decompression overhead. We present the details of decompression algorithm in [1]. The above analysis completes the proof of Lemma 6.

### 5.2 Anchored Graph Compression

We next introduce our second variant of SPGC, notably, *anchored* SPGC, which permits a decompression-free, inference friendly compression, *relative* to a specific set of nodes of interests.

We present our main result below.

**Theorem 7:** Given $\mathcal{M}^L$ and $G$ with a set of targeted nodes $V_A$, there exists an IFGC that computes a compressed graph in $O(|G_L|)$ time to preserve the inference output for every node in $V_A$ at an inference cost in $O(L|E_c|F + L|V_c|F^2)$ time. Here $|G_L|$ refers to the subgraph of $G$ induced by $L$-hop of $V_A$, and $|V_c|$ and $|E_c|$ are bounded by $|G_L|$. $\qquad\square$

As a constructive proof, we introduce a notion of anchored relation, and construct such an IFGC. Given a graph $G$ with a set of designated targeted nodes $V_A$, and a class of GNNs $\mathcal{M}^L$, a graph compression scheme $(C, \mathcal{P})$ is an IFGC relative to $V_A$, if (1) $|\mathcal{P}(C(G))| < |G|$; and (2) for any GNN $M \in \mathcal{M}^L$, and any $v \in V_A$, $M(G, \{v\}) = M(\mathcal{P}(C(G), \{v\})$.

**Anchored relation**. Given graph $G$, an integer $L$, and a designated *anchor* set $V_A \subseteq V$, we define the $L$-hop neighbors of $V_A$, denoted as $N_L(V_A)$, as $\bigcup_{v \in V_A} N_L(v)$, where $N_L(v)$ refers to the set of nodes within $L$-hop of $v$ in $G$. An *anchored relation* $R_L^A$ *w.r.t.* $V_A$ refers to the structural equivalence relation defined over the subgraph $G_L$ of $G$ induced by $N_L(V_A)$.

One may verify that (1) $R_L^A$ is an equivalence relation over $V$, and (2) $R^S = R_L^A$ if $V_A = V$, and $L$ is larger than the diameter of $G$.

**Anchored** SPGC. Given graph $G$ and an anchor set $V_A$ from $G$, an anchored SPGC, denoted as ASPGC, is a graph compression where $C$ computes a compressed graph that is the quotient graph of $R_L^A$.

The computation of compressed graph $G_c$ using ASPGC follows its counterpart in SPGC. The difference is that it induces a subgraph $G_L$ of $G$ with the $L$-hop neighbors of all the nodes in $V_A$. It then invokes the compression algorithm of SPGC to derive $R_L^A$ and applies the compression algorithm of SPGC on $G_L$ to compute the $G_c$ and memoization structure $\mathcal{T}$. The inference process over $G_c$, similarly, follows its SPGC counterpart over $G_c$, which consistently leverages $\mathcal{T}$ to efficiently recover the auxiliary information of neighborhoods. Hence, it preserves the inference results of the nodes in $V_A$ for GNNs classes with up to $L$ layers. In practice, one may set $V_A$ simply as a set of test nodes $V_T$ to adapt ASPGC for specific inference queries. We present details and an example in [1].

**Analysis**. The correctness of ASPGC follows from the data locality of $L$-layered GNN inference when $V_A$ is specified, which only involves the subgraph $G^L$ of $G$ induced by $L$-hop neighbors of $V_A$. ASPGC next follows SPGC to correctly compute $G_c$ from $G_L$. For compression cost, it takes $O(|N^L(V_A)| + |E|)$ time to induce $G_L$, and $O(|V_L| + |E_L|)$ time to construct $G_c$ from $G_L$. (3) The inference cost is consistently $O(L|E_c|F + L|V_c|F^2)$, where both $|E_c|$ and $|V_c|$ are bounded by $|G_L|$.

Given the above analysis, Theorem 7 follows.

## 6 EXPERIMENTAL STUDY

Using both real-world graph datasets and large synthetic graphs, we conducted four sets of experiments, to understand (1) effectiveness of our compression methods, in terms of compression ratio, and the trade-off between inference cost and accuracy loss; (2) their efficiency, in terms of the compression cost and inference cost, (3) impact of critical factors, and (4) an ablation study to evaluate the effectiveness of memoization, and decompression overhead.

### 6.1 Experimental Settings

**Datasets**. We employ four real-world graph benchmark datasets (summarized in Table. 5): (1) **Cora** [40], a citation network where nodes represent documents, and edges are citations among the documents. Node features are described by a 0/1-valued word vector indicating the absence/presence of the corresponding word from the dictionary; (2) **Arxiv** [28], a citation network with nodes representing arXiv papers and edges denoting one paper cites another one. The features of each node includes a 128-dimensional feature vector obtained by averaging the embeddings of words in its title and abstract; (3) **Yelp** [54] is prepared from the raw json data of businesses, comprising a dense network of user-business interactions. Nodes represent users and edges are generated if two users are friends. Node features are the word embeddings derived by reviews of products; (4) **Products** [28], a product co-purchase network, where nodes represent products sold in Amazon and edges denote products purchased together; and (5) **ogbn-papers100M**, a billion-scale citation graph of 111 million papers and 1.6 billion edges. Its graph structure and node features are constructed in the same way as **Arxiv**. The size of test nodes $|V_T| = 5\% * |V|$ across all datasets. For ASPGC, $|V_A| = |V_T|$ as $V_A = V_T$ by default.

Besides real-world benchmark datasets, we also generated a large synthetic dataset **WS-MAG**, by extending a core of real MAG240M network [27] (a citation network) with a small world generator [50].

| Dataset | \|V\| | \|E\| | # node types | # attributes |
|---------|-----|-----|-------------|-------------|
| Cora | 2,708 | 5,429 | 7 | 1,433 |
| Arxiv | 169K | 1.2M | 40 | 128 |
| Yelp | 717K | 7.9M | 100 | 300 |
| Products | 2.4M | 61.9M | 47 | 100 |
| WS-MAG | 2M | 1.192B | 153 | 768 |
| ogbn-Papers100M | 111M | 1.6B | 40 | 128 |

**Table 5: Summary of Datasets.**

**WS-MAG** has a fixed number of nodes (|V| = 2 million), while the number of edges |E| increases linearly from 4 million to 1.192 billion.

**Graph Neural Networks**. We have pre-trained three classes of representative GNNs: GCNs[31], GATs[49], and GraphSAGE[24], for each dataset. For a fair comparison, (1) for all the datasets, we consider node classification, and (2) all compression methods are applied for the same set of GNNs.

**Compression Methods**. We compare SPGC and its variants, $(\alpha, r)$-SPGC and ASPGC, with two state-of-the-art compression methods: (1) DSpar [37], a graph sparsification method that performs edge down-sampling to preserve graph spectral information, and (2) FGC [32], a latest learning-based graph coarsening approach that learns a coarsened graph matrix and feature matrix to preserve desired graph properties, such as homophily. We are aware of other learning-based approaches, yet they are model-specific and require the learned model parameters. Our work is orthogonal to these methods, and is not directly comparable. When conducing the experiments using DSpar and FGC, we set the longest waiting limit = 5 hours which means if the compression graph cannot be generated after 5 hours. We simply exclude them since the compression cost is already comparable to training a new GNNs model.

**Evaluation Metrics**. Given graph $G$, a class of GNNs $\mathcal{M}^L$, a graph compression scheme $(C, \mathcal{P})$ that computes a compressed graph $G_c$, and its matching inference process over $G_c$, we use the following metrics. (1) For efficiency, we evaluate (a) the time cost of compression, and (b) the speed-up of inference, which is defined by $\frac{T_{MG}}{T_{MC}}$, where $T_{MG}$ refers to the inference time cost over $G$, and $T_{MC}$ represents its counterpart over $G_c$. (2) For effectiveness, we report (a) a normalized compression ratio, which is defined as ncr = $1 - \frac{|G_c|}{|G|}$; Intuitively, it quantifies the fraction of $G$ that is "reduced": the larger, the better; and (b) the model performance quantified by accuracy and $F1$-score over $G_c$. In particular for Yelp, over which the benchmark task is a multi-class node classification, we report micro $F1$-score. We report the average performance of 200 inference tests, for each GNN model over each dataset.

**Environment**. SPGC and its variants are developed in Python with PyTorch Geometric [20] and BisPy [3] libraries. All tests are conducted on 4 Intel(R) Xeon(R) Silver 4216 CPU @ 2.10GHz, 128 GB Memeory, 16 cores, and 1 32GB NVIDIA V100 GPU. Our source code, datasets, and a full version of the paper are made available[3].

### 6.2 Experimental Results

**Exp-1: Effectiveness: Accuracy vs. Speed-up.** Fig. 7 compares $(0.5, 1)$-SPGC with DSpar and FGC, in terms of inference speed-up in left figure and inference accuracy/F1-score in the right figure, over all four real datasets. Note that FGC can only generate results for **Cora** since its compression cannot be completed within 5-hours for all other datasets. Here a test annotated as "compression method-GNN" refers to the setting that a GNN inference is applied on a compressed graph generated by the method. (1) $(0.5, 1)$-SPGC outperforms DSpar and FGC across all four datasets for all the GNN classes on inference speed-up. It can improve the inference efficiency better over larger graphs. For example, for **Arxiv**, $(0.5, 1)$-SPGC achieves a speed-up of 3.4 for inference with GraphSAGE, while DSpar achieves a speed-up to 1.5. (2) Consistently, we found that SPGC achieves higher ncr than DSpar and FGC. For example, for **Cora**, SPGC achieves ncr up to 74.5% while DSpar and FGC achieves 46.2% and 30.7% respectively. (3) SPGC outperforms DSpar and FGC in terms of its ability to preserve inference results for GNNs. We observe that i) $(0.5, 1)$-SPGC achieves inference accuracy comparable to direct inference on the original graph across all four datasets (as illustrated in Fig. 7); and ii) $(0.5, 1)$-SPGC retains highest inference accuracy/F1-scores across all datasets and models. For example, for **Cora**, $(0.5, 1)$-SPGC achieves 0.71 accuracy which outperforms 0.68 and 0.67 achieved by DSpar and FGC.

---
[3]https://github.com/Yangxin666/SPGC

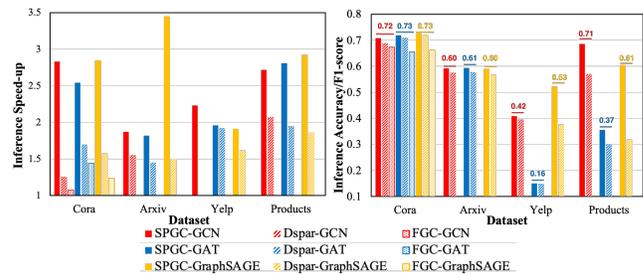

**Figure 7: Comparison of** $(0.5, 1)$-SPGC **with the Baselines in Inference Speed-up and Inference Accuracy/F1-score. (the inference accuracy of** ASPGC **marked in colored lines,** *e.g.,* **the red lines show inference accuracy of** ASPGC **using** GCN; **the inference accuracy on original graph** $G$ **marked in corresponding colors for differernce** GNNs, *e.g.,* **the blue 0.73 indicates the inference accuracy of GAT on original** $G$ **of Cora).**

**Exp-2: Effectiveness: Impact of Factors**. We first investigate the impact of number of layers $L$, which evaluates whether the quality of SPGC-based compression is affected by the complexity of GNNs classes. Then we evaluate the performance of configurable compression $(\alpha, r)$-SPGC, in terms of the impact of $\alpha$ and $r$.

*Varying Number of Layers*. We varied the number of layers of GNNs from 2 to 4 over **Arxiv** and report its impact on inference speed-up (resp. accuracy) in Fig. 8(a) (resp. 8(b)). (1) $(0.5, 1)$-SPGC consistently outperforms all the baselines in both inference speed-up and accuracy, due to that it preserves the inference results with small compressed graphs. (2) In general, the speed-up of inference achieved by $(0.5, 1)$-SPGC is not sensitive to the number of layers. This verifies our theoretical analysis that it preserves inference results with unique smallest compressed graphs, which is independent of model complexity. (3) While the accuracy of GNNs drops as the number of layers become larger, in all cases, $(0.5, 1)$-SPGC preserves the accuracy with smallest "gap" compared with other methods, for the same class of GNNs.

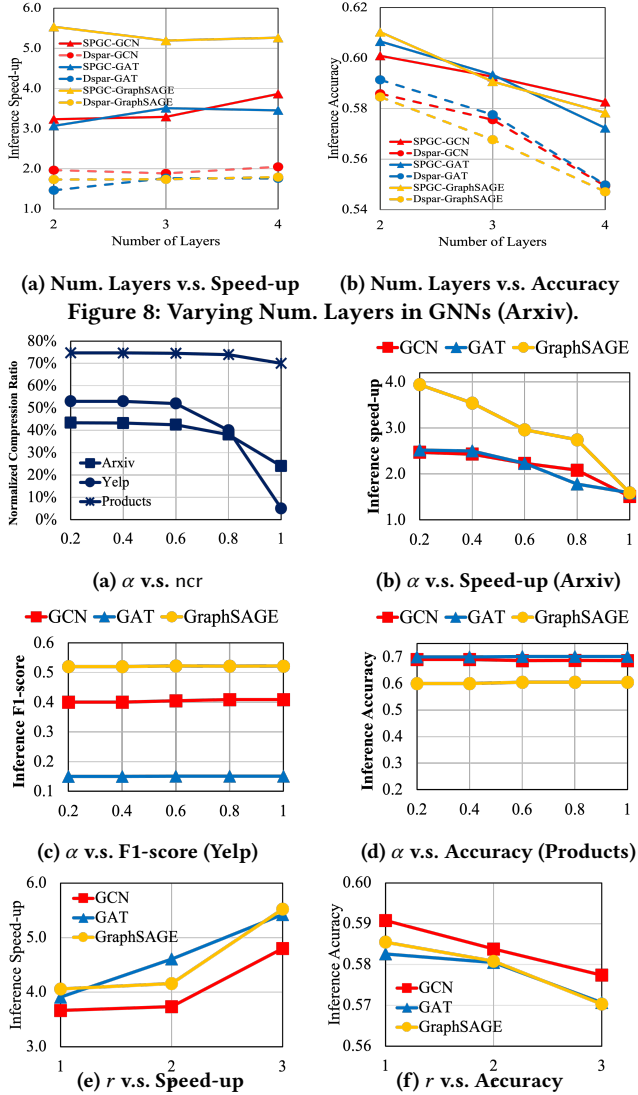

**(a)** Num. Layers v.s. Speed-up    **(b)** Num. Layers v.s. Accuracy

**Figure 8: Varying Num. Layers in GNNs (Arxiv).**

**(a)** $\alpha$ v.s. ncr

**(b)** $\alpha$ v.s. Speed-up (Arxiv)

**(c)** $\alpha$ v.s. F1-score (Yelp)

**(d)** $\alpha$ v.s. Accuracy (Products)

**(e)** $r$ v.s. Speed-up

**(f)** $r$ v.s. Accuracy

**Figure 9: Varying $\alpha$ and $r$ in $(\alpha, r)$-SPGC**

<u>Varying $\alpha$.</u> We varied $\alpha$ from 0.2 to 1, and report the results in Figs. 9(a) to 9(d). It tells us the followings.

(1) As $\alpha$ is increased from 0.2 to 1, ncr drops as illustrated in Fig. 9(a). As expected, larger $\alpha$ makes it harder for $(\alpha, r)$-SPGC to merge nodes that are less close in their representations, leaving more nodes in compressed graphs, hence worsening compression ratio. (2) Consistently for larger $\alpha$ and for all the three GNNs classes, it is harder for $(\alpha, r)$-SPGC to improve their inference efficiency due to larger compressed structure $G_c$ as shown in Fig. 9(b). We observe consistent results for Yelp and Products (not shown; see [1]).

(3) As $\alpha$ increases, the F1-score (resp. accuracy) of the inference results over **Yelp** (resp. **Products**) remains insensitive, as shown in Figs.9(c) (resp. 9(d)). Our observation over **Arxiv** remains consistent, and we omitted it due to limited space. This indicates that $(\alpha, r)$-SPGC does not lose much on the quality of the inference while significantly improved inference efficiency.

<u>Varying $r$ in SPGC.</u> Fixing $\alpha = 0.25$, we vary $r$ from 1 to 3 over **Arxiv** and report the result in Fig. 9. (1) As $r$ is varied from 1 to 3, the

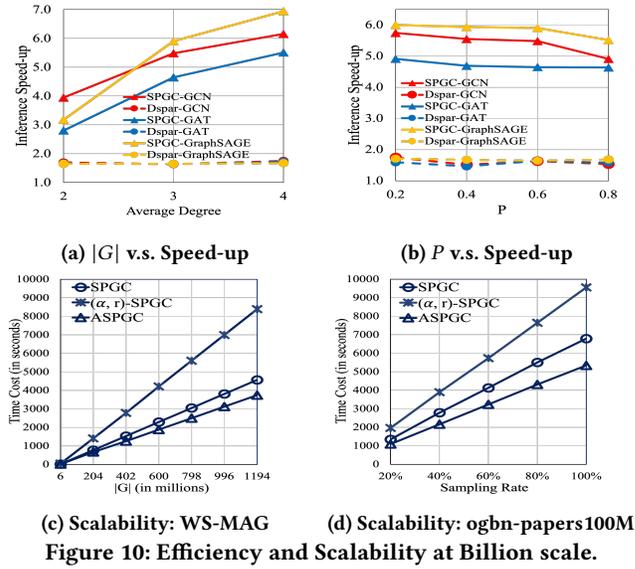

**(a)** $|G|$ v.s. Speed-up    **(b)** $P$ v.s. Speed-up

**(c)** Scalability: WS-MAG    **(d)** Scalability: ogbn-papers100M

**Figure 10: Efficiency and Scalability at Billion scale.**

inference speed-up achieved by $(0.25, r)$-SPGC for all GNNs classes notably increased. Indeed, larger $r$ allows $(\alpha, r)$-SPGC to find and merge more node pairs with equivalent embeddings, which may not be direct neighbors of another pair in the $(\alpha, 1)$-relation. (2) As $r$ increases, the inference accuracy for all GNNs classes slightly drops, and all within a small range of 0.02. This demonstrates that $(\alpha, r)$-SPGC is capable of preserving inference accuracy while increasing $r$ in trading for larger speed up.

**Exp-3: Efficiency & Scalability.** We next evaluate the compression and inference costs of SPGC and its variants.

<u>Average degree and "small-world" effect v.s. Inference Speed-up.</u> We simulate **WS-MAG** based on $(K, P)$ Watts-Strogatz algorithm [50] with fixed $|V| = 2M$ from **MAG240M** dataset. $K$ and $P$ represent average degree and re-wiring probability respectively. As $P$ goes up, the less "small-world" (*i.e.*, more random) the graphs become.

Fixing $P = 0.6$, we vary the average degree from 2 to 4. Fig. 10(a) shows that the inference speed-up for different GNNs types. (1) As the average degree of the graph increases from 2 to 4, inference speed-up achieved by SPGC increases approximately linearly from 3.0×-4.0× to 5.0×-7.0×. (2) Inference speed-up achieved by DSpar remains relatively stable (1.6×-1.8×) and smaller than SPGC. As the average degree goes up, SPGC may takes the advantage of higher density that makes nodes more likely to be merged, resulting in a smaller $G_c$ and greater speed-up.

Next, fixing $|G| = 8M$, we increase $P$ from 0.2 to 0.8. We have the following observations. (1) As $P$ increases, the inference speed-up achieved by SPGC on three GNNs exhibits a slight drop, albeit still much larger than the speed-up by DSpar. (2) Inference speed-ups achieved by DSpar on three GNNs stay flat without even breaking 2× speed-up. This indicates that SPGC may perform better to compress "small-world" graphs, which are common in real-world networks such as social networks [7, 50].

<u>Scalability Tests.</u> We evaluated the scalability of SPGC, $(\alpha, r)$-SPGC and ASPGC on large-scale graphs up to the billion-scale using **WS-MAG** (see Section 6.1) and **ogbn-papers100M** [28].

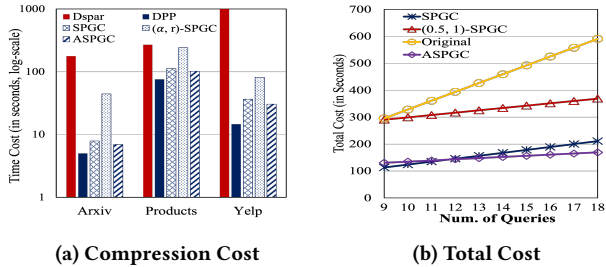

**(a) Compression Cost**      **(b) Total Cost**

**Figure 11: Time Cost Analysis (left: compression cost comparison; right: total cost for varying query loads).**

We varied the size of **WS-MAG** by enlarging its sizes with a generator that uses Watts-Strogatz algorithm [50]. We varied the graph size $|G|$ from 6 million to 1.194 billion in a sequence. As shown in Fig. 10(c), We observe the following. (1) SPGC, ASPGC, and $(\alpha, r)$-SPGC scale well with $|G|$. This is consistent with our time cost analysis on the one-time compression in Table. 1. (2) $(\alpha, r)$-SPGC incurs a relatively higher and more sensitive compression cost due to the overhead of computing and refining the $(\alpha, r)$-relation. The overall cost remains to grow linearly with graph size, while its additional overhead yields improved compression ratios, leading to higher inference speed-ups on the compressed graphs $G_c$.

We also conducted a scalability test using a real-world billion-scale **ogbn-papers100M** (as illustrated in Fig. 10(d)). We generate subgraphs from 20% to 100% of the size of **ogbn-papers100M** using a subgraph sampling approach [17] and evaluate the compression costs of SPGC and its variants. As the sampling rate increases, we observe that SPGC, ASPGC, and $(\alpha, r)$-SPGC all scale well. $(\alpha, r)$-SPGC incurs relatively higher and more sensitive compression costs, consistent with its performances on **WS-MAG**.

*Compression Cost Comparison.* We compare the one-time compression cost induced by the SPGC or its variants with the cost of DSpar as shown in Fig. 11(a). Note that the compression cost of SPGC and its variants include the time cost of DPP. We have the following discoveries. (1) SPGC, $(\alpha, r)$-SPGC and ASPGC outperforms DSpar on all the real-world datasets. In particular, SPGC are 95.52% and 58.45% times faster than DSpar over **Arxiv** and **Products**, respectively. For **Yelp**, DSpar does not run to completion after 1,000 seconds. (2) For SPGC and its variants, ASPGC outperforms SPGC and $(\alpha, r)$-SPGC by 12.71% and 68.34% on average. This is because ASPGC exploits data locality of GNN inference and perform compression up to certain hops of the anchored nodes.

*Total cost: Varying query workloads.* We compare the total costs of SPGC, $(\alpha, r)$-SPGC, ASPGC to the total cost induced by the original $G$. For SPGC and its variants, the total cost is the sum of "one-time" compression cost and the inference cost for all inference queries; for "Original", it refers to the inference time on the original graph. Varying the number of inference queries from 9 to 18 (each with fixed number of test node $|V_T| = 0.05|V|$), Fig. 11(b) reports the total cost of a 3-layers GCN on **Products**. (1) All SPGC methods are able to reduce the original inference cost, and all can improve the inference efficiency better with larger amount of queries. (2) Among SPGC variants, for queries with given anchored nodes, ASPGC performs best in improving the inference efficiency. We also observe that $(\alpha, r)$-SPGC is more sensitive to different datasets, compared with SPGC and ASPGC. Indeed, SPGC enforces rigidly embedding

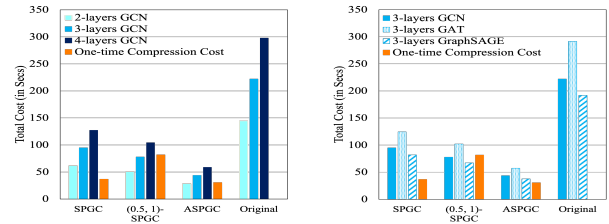

**(a) Varying** GNN **Layer Number**    **(b) Varying** GNN **class**

**Figure 12: Total Cost: Varying** GNN **Layers and Class.**

equivalence, while ASPGC benefits most from data locality from anchored node set of fixed size, hence both are less sensitive to datasets. $(\alpha, r)$-SPGC on the other hand is most adaptive for tunable trade-off between inference speed up and compression ratio, depending on the heterogeneity of node embeddings.

*Total cost: varying* GNN *complexity.* We next investigate the impact of GNN model complexity, in terms of number of layers, and types. Using GCNs, we varying the number of layers $L$ from 2 to 4 (We denote $GCN_2$ as the 2-layers GCN; similarly for others), and a query workload of 3 different inference queries with the same size $|V_T|$. Fig. 12(a) tells us that as the number of layers increase, the inference cost grows as expected. ASPGC outperforms SPGC and $(0.5, 1)$-SPGC consistently for all layers in compression efficiency.

Fixing query workload size as 3, and the number of layers as 3 for all GNN types, we report the total cost over GCN, GAT and GraphSAGE. As shown in Fig. 12(b), all three types of GNNs consistently and significantly benefit from SPGC, with an improvement of inference time by a factor of 2.21, 2.36 and 2.11, respectively; similar for $(\alpha, r)$-SPGC and ASPGC. The one-time compression costs are consistent with our prior observations in Fig. 12(a). In general, GAT may benefit most from inference-friendly compression. A possible reason is that the memoization effectively cached its additional edge weights as part of the auxiliary information, which are a source of overhead for inference over $G$.

We have also investigated (1) how memoization structure $\mathcal{T}$, and neighbor recovery affect the inference accuracy and speed-up achieved by SPGC; (2) the impact of aggregation method and improvement of model training cost. Due to limited space, we present additional tests with details in [1].

## 7 CONCLUSION

We have proposed IFGC, an inference friendly graph compression scheme to generate compressed graphs that can be directly processed by GNN inference process to obtain optimal inference output. We have introduced a sufficient condition, and introduced three practical specifications of IFGC, SPGC, for inference without decompression, and configurable $(\alpha, r)$-compression, to achieve better compression ratio and reduction of inference cost, and anchored SPGC, that preserves inference results for specified node set. Our theoretical analysis and experimental study have verified that IFGC-based approaches can significantly accelerate the inference on real-world and large graphs with small loss on inference accuracy. A future topic is to extend our compression scheme to accelerate GNN training. Another topic is to develop parallel compression scheme for large graphs.

# 8 APPENDIX

## 8.1 Appendix A: Proofs

**Proof of Lemma 1**. Given $\mathbb{M}$ and $G$, the binary relation $R_M^L$ is an equivalence relation, *i.e.*, it is reflexive, symmetric, and transitive.

PROOF. (1) It is easy to verify that $R_M^L$ is reflexive, *i.e.*, $(v, v) \in R_M^L$ for any $v \in V$: for any node $v$, and any of its neighbor $v'$, an identity mapping verifies the reflexiveness of $R_M^L$. (2) We show the symmetric by definition. If $(v, v') \in R_M^L$, then there is a "matching relation" $h$, where $h(v)=v'$, such that for every neighbor $u$ of $v$, there is a neighbor $u' = h(u)$ such that $(u, h(u)) \in R_M^L$. Consider the inverse relation $h^{-1}$, we can verify that $(h(u), h^{-1}(h(u))) = (h(u), u) = (u', u) \in R^S$. This holds for every neighbor $u'$ of $v'$, which leads to that $(v', v) \in R^S$. (3) The transitivity of $R_M^L$ states that for any pair $(v_1, v_2)$ and $(v_2, v_3)$ from $G$, if $(v_1, v_2) \in R_M^L$ and $(v_2, v_3) \in R_M^L$, then $(v_1, v_3) \in R_M^L$. Let $h$ be the matching relation that induces $R_M^L$, where $h(v_1) = v_2$, and $h(v_2) = v_3$. Then consider a composite relation $h' = h \circ h$. We have $h'(v_1) = h(h(v_1)) = v_3$. Similarly, we can verify that $h'^{-1}(v_3) = v_1$. Hence $(v_1, v_3) \in R_M^L$ as induced by $h' = h \circ h$. The transitivity of $R_M^L$ follows. □

**Proof of Lemma 2.** Given a class of GNNs $\mathbb{M}^L$ and a graph $G$, there exists a IFGC $(C, \mathcal{P})$ *w.r.t.* $\mathbb{M}^L$ and $G$, if there is a nontrivial inference equivalent relation $R_M^L$ *w.r.t.* $\mathbb{M}^L$ (1) $C(G)$ computes a quotient graph $G_c$ induced by a nontrivial inference equivalent relation $R_M^L$ *w.r.t.* $\mathbb{M}^L$ and $G$, and (2) $|\mathcal{P}(G_c)| < |G|$.

PROOF. Let $R_M^L$ be a non-empty inference equivalent relation *w.r.t.* $\mathbb{M}^L$ and $G$, and $G_c$ be the quotient graph induced by $\mathbb{M}^L$. (1) Given Lemma 1, $R_M^L$ is a nontrivial equivalence relation. As $R_M^L \neq \emptyset$, there exists at least one equivalent class $[v]$ with size larger than one, *i.e.*, $|G_c| < |G|$. (2) As the post processing function $\mathcal{P}$ over $G_c$ does not introduce additional nodes or edges, as ensured by $|\mathcal{P}(G_c)| < |G|$, we have $|\mathcal{P}(C(G))| < |G|$.

We next show that $M(G) = M(\mathcal{P}(C(G)))$. To see this, it suffices to show that for every node $v \in G$, $M(G, v) = M(\mathcal{P}(C(G)), [v])$. Indeed, if this holds, the output representation of $M(G_c, [v])$ can be readily assigned to every node $v \in [v]$ without decompression. Further, it suffices to show that for any nodes $(v_i, v_j) \in [v]$, $M(v_i, G) = M(v_j, G) = M([v], G_c)$.

We prove the above result by conducting an induction on the layers $k$ of the GNNs. The general intuition is to show that it suffices for the compression function $C$ to track some auxiliary information of the neighbors of $v$ in $G$ when generating the compressed graph $G_c$, such that the aggregation result of the node update function $X_v$ for each node $v$ can be readily computed by a weighted aggregation in $G_c$ in a post processing function $\mathcal{P}$, *without* decompression (hence incurs no additional inference cost). Such information can be readily tracked and looked up as needed at inference time over $G_c$, by storing $R_M^L$ equivalently as a "matching" relation $h$ between each node $v$ and its equivalence class $[v]$.

Given any two nodes $v$ and $v'$ such that $(v, v') \in R_M^L$, *i.e.*, $v \sim_M^L v'$, consider an equivalent characterization of $R_M^L$ as a mapping $h$ between $G$ and $G_c$ such that $h(v) = [v]$ in $G_c$, for each node $v \in G$. (1) Let $k = 0$. Then one can verify that $M(v_i, G) = X_{v_i}^0 = X_{v_j}^0 = M^0(v_j) = M^0([v]) = M([v], G_c)$.

(2) Let $k = i$, and assume the result holds for any GNNs in $\mathbb{M}^L$ at layer $k = i$. That is, for every node $v \in G$, $M^i(v_i, G) = M^i(\mathcal{P}(C(G)), [v])$. As $[v]$ is an equivalence class induced by $R_M^L$, $M^i(v, G) = M^i(v', G) = M^i([v], G_c)$ for $i$-layered GNNs $\mathbb{M}^i$ with the same node update function $M_i$. Consider the output of $M^{i+1}(v, G)$. The computation invokes the node update function as $X_v^{i+1} = M_v(\Theta^i, \text{AGG}, N(v), M_v^i)$, and $X_{[v]}^{i+1} = M_{[v]}(\Theta^i, \text{AGG}, N([v]), M_{[v]}^i)$. As we consider fixed, deterministic GNNs with the same node update function,

- the model weights $\Theta^{i+1} = \Theta^i$,
- operator AGG remain to be fixed,
- $M_v^i = M_{[v]}^i$ by induction; and
- the matching relation $h$ ensures the invariant that for *every* neighbor $u \in N(v)$, there exists a counterpart $[u'] \in N(h(v)) = N([v])$ in the quotient graph $G_c$, such that $X_u^i = X_{[u]}^i$, ensured by $R_M^i$ and the definition of quotient graphs.

We now show that $X_v^{k+1} = M_v(\Theta^k, \text{AGG}, N(v), M_v^k) = M_{[v]}(\Theta^k, \text{AGG}, N([v]), M_{[v]}^k) = X_{[v]}^{k+1}$. To see this, $M_{[v]}$ only needs to "invoke" a fast look up function $\mathcal{P}$ that retrieves an edge weight (pre-stored during compression $C$; see "Notes") to adjust its direct aggregation over $N_{[v]}$ in $G_c$, as needed, without decompression. We illustrate below typical examples of the weighted update for major GNNs in Table 4 augmenting Table. 3; where the weights are highlighted in bold and red.

Hence, for any pair of nodes $(v, v') \in [v]$, $M(v, G) = M(v', G) = M([v], G_c)$ for any GNN $M \in \mathcal{M}^L$. By definition, the graph computation scheme $(C, \mathcal{P})$ is an IFGC for $\mathcal{M}^L$ and $G$. □

**Remark**. The above analysis verifies that $G_c$ *preserves* the inference results, by showing that the computation of $M_v$ in $G$ can be simulated by an equivalent, re-weighted inference-friendly counterpart $M_{[v]}$ that directly process $G_c$ without decompression. The weights can be easily bookkept during compression, tracked by a *run-time* look-up function $\mathcal{P}$ along with the inference process, over a (smaller) $G_c$, without a stacked run of $\mathcal{P}$, without incurring additional time cost, and without decompression. Moreover, this incurs only a small bounded memory cost of up to $|E|$ weights (numbers). We refer the implementation details in Sections 4, 5.1 and 5.2, respectively, for IFGC specifications.

**Proof of Theorem 3**. Given a class of GNN $\mathbb{M}^L$ and graph $G$, the relation $R^S$ over $G$ is an inference equivalence relation *w.r.t.* $\mathbb{M}^L$.

PROOF. We first show that $R^S$ is an equivalence relation, *i.e.*, it is reflexive, symmetric, and transitive. (1) It is easy to verify that $R^S$ is reflexive and symmetric, *i.e.*, $(v, v) \in R^S$ for all the nodes in $G$, and for any node pair $(v, v') \in R^S$, $(v', v) \in R^S$, by definition. (2) To see the transitivity, let $(v_1, v_2) \in R^S$, and $(v_2, v_3) \in R^S$. Consider a matching relation $h$ such that for each pair $(v, v') \in R^S$, $h(v) = v'$. Then $h(v_1) = v_2$, $h(v_2) = v_3$. We can verify that $h(v_1) = v_3$, which induces that $(v_1, v_3) \in R^S$, by the definition of structural equivalence. Hence $R^S$ is an equivalence relation.

We next show that $R^S$ is an inference equivalence relation *w.r.t.* $\mathcal{M}^L$, that is, for any pair $(v, v') \in R^S$, $v \sim^L_{\mathcal{M}} v'$. Similarly as the analysis for Theorem 2, we perform an induction on the number of layers $k$ of GNNs $M$.

(1) Let $k = 0$. Then $M(v, G) = X^0_{v_i} = X^0_{v_j} = M^0(v_j) = M^0([v]) = M([v], G_c)$.

(2) Assume $R^S$ is an inference equivalence relation *w.r.t.* $\mathcal{M}^L$ for $k = i$. Given $(v, v') \in R^S$, for the layer $i + 1$, $X^{k+1}_v = M_v(\Theta^k, \text{AGG}, N(v), M^k_v) = M_{[v]}(\Theta^k, \text{AGG}, N([v]), M^k_{[v]}) = X^{k+1}_{[v]}$, following the similar analysis as in $R^L_M$ counterpart. Note that the $h$ matching function of $R^S$ is specified for $R^S$; and $\mathcal{P}$ is an identity function. □

**Proof of Theorem 4**. Given a class of GNNs $\mathbb{M}^L$ with the same node update function $M_v$, and a graph $G$, a SPGC produces a unique, minimum compressed graph $G_c$, up to graph isomorphism over the quotient graphs induced by $R^S$.

Proof. To see this, we specify the compression process $C$ of SPGC to be a function that computes the *largest* structural-equivalence relation $R^{S*}$ *w.r.t.* $\mathbb{M}^L$ and $G$. We first show that there exists a unique largest inference-friendly relation $R^{S*}$ over $G$. We then show that the unique, largest $R^{S*}$ induces a minimum compressed graph $G^*_c$, up to graph isomorphism over the quotient graphs induced by $R^S$.

(1) We prove the uniqueness property by contradiction. Assume there are two largest structural equivalence relations $R^{S*}$ and $R'^{S*}$, where $|R^S| = |R'^{S*}|$, and $R^{S*} \neq R'^{S*}$. Let $(v, v') \in R^{S*}$, and $(v, v') \neq R'^{S*}$. For the latter case, either $X^0_v \neq X^0_{v'}$, or there exists a neighbor $u \in N(v)$ such that there is no neighbor $u'$ in $N(v')$ such that $(u, u') \in R'^{S*}$. For the first case, clearly $(v, v') \neq R^{S*}$. For the second case, there exists a GNN $M \in \mathcal{M}^L$ such that $M(u, G) \neq M(u', G)$, as $u'$ ranges over all the neighbors of $v'$. This indicates that $(v, v') \neq R^S$, which contradicts to that $R^S$ is an inference equivalence relation. Hence $R^{S*} = R'^{S*}$.

(2) We consider the notion *graph isomorphism* defined over compressed graph. We say two compressed graphs $G_c$ and $G'_c$ are isomorphic, if there exists a bijective function $h_c$ between $G_c$ and $G'_c$, such that for any edge $([u], [v])$ in $G_c$, $(h_c([u]), h_c([v]))$ in $G'_c$.

Given that $R^{S*}$ is the unique largest structural equivalence relation that is also an inference equivalence relation, let $G^*_c$ be the corresponding quotient graph of $R^{S*}$. Assume there exists another quotient graph $G'^*_c$ induced from $R^{S*}$ that is not isomorphic to $R^{S*}$. Then there exists at least an edge $([u], [v])$ in $G^*_c$ for which no edge exists in $G'^*_c$. Given that $G^*_c$ is the quotient graph induced by the maximum structural relation $R^{S*}$, then either $G'^*_c$ has a missing edge, which contradicts to that it is a quotient graph induced by the largest $R^{S*}$, or $R^{S*}$ is not inference equivalence, which contradicts to that it is a structural equivalence relation, given Theorem 3. Hence there exists a unique, smallest quotient graph induced by $R^{S*}$ up to graph isomorphism. □

**Discussion of Compression Ratio by** SPGC. Given GNNs $\mathbb{M}^L$ and graph $G$, SPGC achieves (1) an optimal compression ratio $\frac{|G|}{|G_c|}$, where $G_c$ is the unique minimum quotient graph induced by the maximum $R^S$ *w.r.t.* $\mathbb{M}^L$ and $G$; and (2) an optimal speed up of inference for $\mathbb{M}^L$ at $\frac{d|G|}{|G_c|}$ (where $d$ is the maximum degree of $G$), independent of GNN configurations.

Given that there exists a unique smallest compressed graph $G^*_c$ induced by the largest structural equivalence relation $R^S$, a theoretical optimal compression ratio cr can be provided as $\frac{|G|}{|G^*_c|}$.

Accordingly, considering an upperbound of the inference time cost of the mainstream GNNs as summarized in Table 3. A maximum speed up for inference cost can be computed as $\frac{O(LmdF^2 + LnF^2)}{O(Lm'd'F^2 + Ln'F^2)} \leq \frac{md+n}{m'+n'} \leq d\frac{m+n}{m'+n'} = d \cdot \text{cr}$.

Interestingly, this result establishes a simple connection between a "best case" speed up and the theoretical optimal compression ratio, in terms of a single factor $d$ that is the maximum degree of $G$. The intuition is that in the "ideal" case, every neighbor $u$ of a node $v$ in $G$ is pairwise indistinguishable for the inference process, thus are "compressed" into a single node $[u]$, introducing a local inference cost reduction at most $d$ times.

**Proof of Theorem 5**. Given a class of GNNs $\mathbb{M}^L$ with the same node update function $M_v$, and a set of graphs $\mathcal{G}$, for any GNN $\mathcal{M} \in \mathbb{M}^L$, a SPGC $(C, \_)$ computes a unique compressed set $\mathcal{G}_c$, such that for any pair $(G, G') \in \mathcal{G}_\mathcal{M}$, there exists a pair $(G_c, G'_c) \in \mathcal{G}_\mathcal{M}$, *i.e.*, the discrminativeness of $\mathcal{M}$ is preserved by SPGC.

Proof. The uniqueness of the set $\mathcal{G}_c$ can be shown by verifying that each compressed graph $G_c \in \mathcal{G}_c$ is a corresponding unique, smallest compressed graph for an original counterpart $G \in \mathcal{G}$.

Let $(G, G')$ be a pair in $\mathcal{G}_\mathcal{M}$ for a GNN $\mathcal{M} \in \mathcal{M}^L$. Then $M(G) = M(G')$. Given that $G_c$ is the unique smallest compressed graph of $G$ induced by an inference-equivalence relation, $M(v, G)$ can be computed via a weighted inference process $M([v], G_c)$, for every node $v$ in $G$. Hence $M(G) = M(\mathcal{P}(G_c))$. Similarly, $M(G') = M(\mathcal{P}(G'_c))$. Thus $M(\mathcal{P}(G_c)) = M(\mathcal{P}(G'_c))$. Given that $\mathcal{M}$ remains a fixed model, $M(G'_c) = M(G_c)$ for each node $[v]$ in $G_c$, hence $(G'_c, G_c) \in \mathcal{G}_{c\mathcal{M}}$. □

## 8.2 Appendix B: Algorithms

**Procedure** DPP. Given $G$, procedure DPP computes the equivalence relation $R$ that satisfy for any node pair $(v, v')$ in $R$, if and only if the followings holds:

- for any neighbor $u$ of $v$ ($u \in N(v)$), there exists a neighbor $u'$ of $v'$ ($u'$ in $N(v')$), such that $(u, u') \in R^S$; and
- for any neighbor $u''$ of $v'$ in $N(v')$, there exists a neighbor $u'''$ of $v$ in $N(v)$, such that $(u'', u''') \in R^S$.

Procedure DPP first computes the rank for all nodes in $G$ and identifies the maximum rank (line 1-3). It next induces the node partition Par based on the rank (line 4). Then it collapses nodes in $B_{-\infty}$ such that only one randomly selected node $v \in B_{-\infty}$ remains and all edges that were incident to the eliminated nodes are redirected to be incident to $v$ (line 5). It next induces the equivalence relation $R_i$ at each rank $i$ (line 6-7). It next prunes each $R_i$ based on whether

---

**Algorithm 3** Procedure DPP($G$)

1: **for** $v \in V$ **do**
2:     compute rank($n$);
3: $\phi := \max\{\text{rank}(n)\}$;
4: node partition Par $:= \{B_i : i = -\infty, 0, \ldots, \phi\}$
5: collapse $B_{-\infty}$;
6: **for** $i = -\infty, 0, \ldots, \phi$ **do**
7:     induce $R_i$ at rank $i$;
8: **for** $n \in V \setminus B_{-\infty}$ **do**
9:     **for** $i = 0, \ldots, \phi$ **do**
10:         $R_i := R_i \setminus \{(v, v') | (v, n) \in E \text{ and } (v', n) \notin E\}$;
11:         update $B_i$;
12: **for** $i = 0, \ldots, \phi$ **do**
13:     $D_i := \{X \in \text{Par} : X \subseteq B_i\}$;
14:     refine $D_i$ with **Paige-Tarjan**;
15:     collapse $X \mid \forall X \in D_i$;
16:     **for** $n \in V \cap B_i$ **do**
17:         **for** $j = i + 1, \ldots, \phi$ **do**
18:             $R_j := R_j \setminus \{(v, v') | (v, n) \in E \text{ and } (v', n) \notin E\}$;
19:             update $B_j$;
20: $R := \bigcup_{i \in \{-\infty, 0, \ldots, \phi\}} R_i$;
21: **return** $R$

---

**Figure 13: Procedure** DPP

---

**Algorithm 4** : $(\alpha, r)$-SPGC

**Input:** Graph $G$, node feature matrix $X$, configuration (xsim, $\alpha, r$); a class of GNNs $\mathbb{M}^L$ with node update function $M$;

**Output:** A compressed graph $G_c$ and $\mathcal{T}'$, $EC$, compressed encodings $\text{AL}_c$ and rules;

1: set $R^{(\alpha, r)} := \emptyset$; set $EC := \{V\}$; set $\mathcal{T}' := \emptyset$; set $AL := \emptyset$; set $\text{AL}_c := \emptyset$; dictionary rules $:= \emptyset$; graph $G_c$, $G_r := \emptyset$;
2: $G_r := (V, E_r) \mid E_r := \{(u, v)\}, v \in V$ and $u \in N_r(v)$;
3: induce adjacency list $AL$ from $E_r$;
4: $R^{(\alpha, r)} := \text{DPP}(G_r)$;
5: $R^{(\alpha, r)} := R^{(\alpha, r)} \setminus \{(v, v') \mid \text{xsim}(X_v^0, X_{v'}^0) < \alpha\}$;
6: $EC := V / R^{(\alpha, r)}$; /* induce partition $EC$ from $R^{(\alpha, r)*}$/
7: $G_c$, $\mathcal{T} := \text{CompressG}(\mathcal{T}, G_c, EC, G, M)$;
8: $\text{AL}_c := \text{Re} - \text{Pair}(\text{AL}_c)$;
9: **return** $G_c$, $\mathcal{T}$, $EC$, $\text{AL}_c$, rules;

---

**Figure 14: Algorithm** $(\alpha, r)$-SPGC

there are edges incident to nodes in $B_{-\infty}$ and updates $B_i$ accordingly (line 8-11). It next iterates over the rank equal to $0, \ldots, \phi$ (line 12). Within each iteration, it conducts the followings: 1) it computes the $D_i$ and refines it using **Paige-Tarjan** and collapses all $X \in D_i$ (line 13-15); 2) it prunes each $R_j$ where $j \in \{i + 1, \ldots, \phi\}$ based on whether there are edges incident to nodes in $B_i$ and updates $B_j$ (line 16-19). It combines all $R_i$ to derive $R$ and returns $R$ (line 20-21).

**Inference process with decompression.** Following the inference algorithm in Sec. 4.4, the users can directly query on $V_T$ from $G_c$ compressed from $(\alpha, r)$-SPGC without any decompression and benefit from the accelerated inference. However, this may result

---

**Algorithm 5** : decompG

**Input:** Compressed graph $G_c$, $G_r$, $EC$, $\text{AL}_c$, rules;
**Output:** A de-compressed graph $G_{cd}$;

1: $G_{cd} = G_c$;
2: sort $\text{AL}_c$ by node degrees in $G_r$;
3: **for** $[v] \in G_c.V_c$ **do**
4:     **while** $\exists$ not visited $v \in [v]$ **do**
5:         $D = \{u | u \in N_r(v) \land \neg \text{ decompressed by } v \text{ in } [v]\}$;
6:         **for** $\forall u \in D$ **do**
7:             $V_{cd}$.add($u$);
8:             $E_{cd}$.add($u$, $[v]$);
9:         mark $v$ as visited;
10: **return** $G_{cd}$;

---

**Figure 15: Algorithm** decompG

in dropped inference accuracy since inference equivalence is not directly preserved. A pair $(v, v')$ in an $(\alpha, r)$-relation $R^{(\alpha, r)}$ is no longer conform to embedding equivalence, thus an $(\alpha, r)$-SPGC $(C, \_)$ alone is not an IFGC, *i.e.*, no longer inference-friendly for a given $G$ and GNNs class $\mathcal{M}^L$. We next show that with a cost-effective decompression process $\mathcal{P}$, a $(1, r)$-SPGC $(C, \mathcal{P})$ becomes inference friendly. The idea is to integrate an inference-time "restoring" of the $r$-hop neighbors up to a local range. We start by introducing an auxiliary structure called *neighbor correction table*. For each node $v \in G$, a neighbor correction table is a compressed encoding of its $r$-hops neighbors $N_r(v)$. There are a host of work on effective encoding of nodes and their neighbors (see [8]). We non-trivially extend Re-Pair compression, a reference encoding method [15, 33] for efficient decompression of $r$-hop neighbors with following two types of pointers that maintain: 1) equivalent class/cluster (nodes within same equivalence relation can be reached from each other) and 2) common $r$-hops neighbors shared by nodes within same equivalence relation.

**Decompression Algorithm.** We next outline decompG shown in Fig. 15 that implements a decompression function $\mathcal{P}$. Upon receiving an inference query defined on $V$, decompG visits every node $v$ in $[v] \in V_c$ exactly once and identifies $D$ such that it captures $N_r(v)$ that have not been decompressed by visited nodes. Then decompG adds new edges between the nodes in $D$ and $[v]$.

**Example 7:** We continue with Example 6. Given $G_{c5}$, $EC$, $\text{AL}_c$, and rules, decompG sorts $\text{AL}_c$ in the descending order of node degrees while still keeping it in the order of equivalent class. decompG next decompresses $N_r([a])$ such that the nodes $b_1 b_2 b_3 c_1$ are decompressed first by $a_1$ and then $b_4 c_2$ are decompressed by $a_2$ as illustrated in the Fig. 16. Similarly, $N_r([b])$ and $N_r([c])$ are decompressed accordingly. decompG returns the decompressed graph $G_{cd}$ as shown in the Fig. 6. □

**Correctness.** decompG recovers the union of up-to r-hop neighbor nodes for all nodes in $[v]$. decompG adds edges such that messages can be passed from up-to r-hop to all nodes in $[v]$ during GNNs inference. This ensures that lost information in compression can be recovered by added neighbor nodes and edges (decompression).

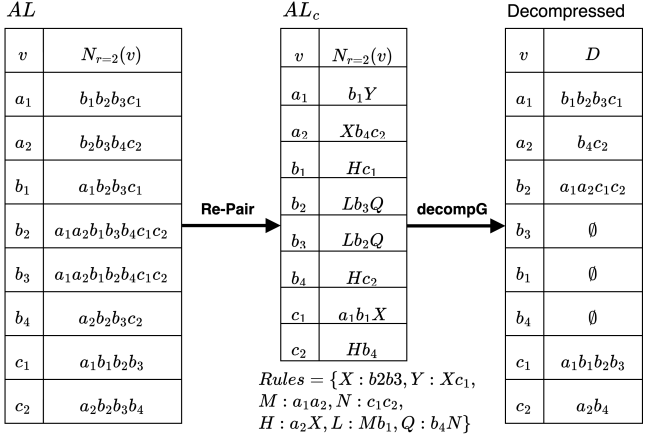

Figure 16: Illustration of the compression by $\mathrm{Re-Pair}$ and decompression by decompG ($G_5$ in Fig. 6 as the example).

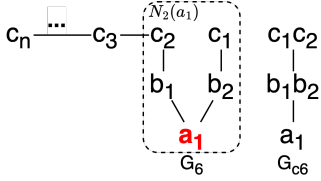

Figure 17: Compressing graph $G_6$ with anchored SPGC: for 2-layered GNNs, with anchored set $V_A = \{a_1\}$.

**Inference Cost.** decompG is in $O(|V|r)$ time. It takes $O(r)$ time to decompress its $r$-hop neighbor nodes of $v$. In the worst case, decompG decompresses $G_c$ back to the size of $G_r$, thus the inference cost on the decompressed graph is in $O(L|E_r|F + L|V|F^2)$ time.

**Lemma 8:** $G_c$ from $(1, 1)$ SPGC $\equiv G_c$ from SPGC. □

*Proof Sketch.* Give $(1, 1)$ SPGC, we have $\alpha = 1$ and $r = 1$. When $r = 1$, the Partition computes the maximum bisimulation derived from $A$, which is same computation as the line 4 in SPGC. When $\alpha = 1$, we assign all nodes in each equivalent class the same labels. Therefore, there are no changes to $EC$ from line 7 to line 11 in $(\alpha, r)$ SPGC. Finally, the line 12 is same as the line 6 in SPGC. Therefore, $G_c$ derived from $(1, 1)$ SPGC is same as the one derived from SPGC.

**An example to illustrate** ASPGC. We consider the following example.

**Example 8:** Consider a class of GNNs $\mathcal{M}^2$, and graph $G_6$ with $V_A = \{a_1\}$ (shown in Fig. 17). For layer 2 GNNs, ASPGC first induces a subgraph $G_6^L$ with 2-hop neighbors of $a_1$. It then follows a SPGC counterpart to compute the compressed graph $G_{c_6}$, with only three nodes. Observe that the compressed graph $G_{c_6}$ does not guarantee to preserve the embedding of other nodes, but only $a_1$. For inference over node $b_1 \notin V_A$, since $G_6^L$ do not cover within 2-hop neighbors of $b_1$, $G_{c_6}$ cannot preserve its embedding. One can further verify that (1) the node pair $(c_1, c_2) \in R_L^A$ but $\notin R^S$, and (2) the anchored compression does not need to consider nodes beyond $L$-hop of anchored nodes (such as the chain from $c_3$ to $c_n$), as it best exploits the data locality of GNN inference process "centered" at $V_A$. □

## 8.3 Appendix C: Additional Experiments

**The impact of aggregation methods on** SPGC. Fixing both $\alpha = 0.25$ and $r = 1$, we select five different aggregation methods: Mean, Sum, Median, Max, and Min used for node embedding construction of node $[v] \in G_c$ to compare the performance of SPGC over ***ogbn-arxiv*** with different aggregation methods. As illustrated in the Table. 7, Mean aggregation achieves the best overall inference accuracy (highest in GCN and GraphSAGE), followed by Median (highest in GAT) while Max and Min fall behind other methods.

**Training Cost and Accuracy of** SPGC. We use **Arxiv** to compare the training cost and accuracy of a 3-layers GCN using the compressed graph $G_c$ from SPGC and $(0.5, 1)$-SPGC to training cost and accuracy using the original graph $G$. Table. 7 reports the training accuracy and time comparison. We observe that, compared to training on the original $G$, SPGC, *i.e.*, SPGC achieves a 40.57% faster training than training on $G$ while retaining comparable accuracy. On the other hand, $(0.5, 1)$-SPGC trades a mere 3.10% loss in accuracy for a 59.96% faster training time compared to the training on $G$. Above results experimentally demonstrate that training on the compressed graph $G_c$ from SPGC and $(\alpha, r)$-SPGC can accelerate the training procedure of GNN while achieving comparable test accuracy compared to the training on original $G$.

**Varying** $\alpha$ **and inference speed-up.** As $\alpha$ increases, it is harder for $(\alpha, r)$-SPGC to improve their inference efficiency due to larger compressed structure. Here we show two additional results from **Yelp** and **Products** datasets in Fig. 18. We observe that as $\alpha$ increases, inference speed-up on **Yelp** and **Products** exhibit similar trends as observed from **Arxiv** datasets.

**Memory Cost Analysis.** We analyze memory consumption by comparing the memory costs associated with the original graph $G$

Table 6: Comparison of Inference Accuracy with Different Aggregation Methods in SPGC (Arxiv).

|        | GCN    | GAT    | GraphSAGE |
|--------|--------|--------|-----------|
| AVG    | **0.5908** | 0.5796 | **0.5855** |
| $\sum$ | 0.5096 | 0.5297 | 0.4602 |
| Median | 0.5901 | **0.5804** | 0.5849 |
| Max    | 0.5766 | 0.5660 | 0.5584 |
| Min    | 0.5758 | 0.5736 | 0.5641 |

Table 7: Comparison of Training Accuracy and Cost of $G_c$ from Original $G$, $(0.5, 1)$ − SPGC, **and** SPGC (Arxiv).

|          | Original $G$ | $(0.5, 1)$ − SPGC | SPGC     |
|----------|--------------|-------------------|----------|
| $|V|$    | 169,343      | 127,508           | 143,939  |
| $|E|$    | 1,166,243    | 205,132           | 607,989  |
| *ncr*    | 0.00%        | 75.09%            | 43.70%   |
| Train Time | 682.63s    | 273.32s           | 405.70s  |
| Train Accuracy | 0.65   | 0.63              | 0.65     |

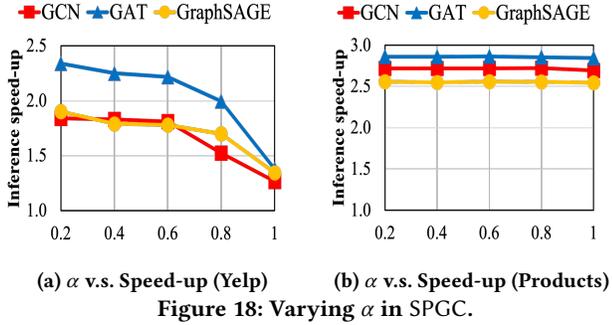

**(a) $\alpha$ v.s. Speed-up (Yelp)**    **(b) $\alpha$ v.s. Speed-up (Products)**

**Figure 18: Varying $\alpha$ in SPGC.**

and its compressed counterpart $G_c$ derived using SPGC. Following the definition of ncr, we define Memory Compression Rate (mcr) as $mcr = 1 - \frac{|M_c|}{|M|}$. It quantifies the fraction of memory cost that is "reduced": the larger, the better; The detailed results are presented in Table 8. To demonstrate the memory savings achieved by $G_c$, we conduct the following three tests:

**(1) Graph Memory (GM) Test**: We measure the graph memory usage as the graph size in the PyTorch Geometric Graph Data format [20] including node features, node labels, and edge index tensors. Comparing to $G$, the GM for $G_c$ is reduced by 38.26%, 21.58%, and 76.01% for **Arxiv**, **Yelp**, and **Ogbn-Products** respectively;

**(2) Memoziation Structure $\mathcal{T}$ Memory (MSM) Test**: We evaluate the memory cost of the memoziation structure $\mathcal{T}$. Our findings indicate that: i) On average, the memory cost of $\mathcal{T}$ accounts for only 10.21% of that of $G_c$ and on average across all the three datasets; and ii) The combined memory cost of $G_c + \mathcal{T}$ still remains significantly smaller than that of $G$, with reductions of 34.56%, 16.72%, and 71.58% for **Arxiv**, **Yelp**, and **Ogbn-Products** respectively;

**(3) Peak Memory (PM) Test**: We measure the peak memory usage during the inference process using a 3-layers GCN model, using a hidden size of 32. Compared to the original graph $G$, the compressed $G_c$ reduces the peak memory by 14.58%, 8.09%, and 65.19% for **Arxiv**, **Yelp**, and **Ogbn-Products**. Notably, as the size of graph increases, the reduction in peak memory becomes more pronounced, reflecting the overall decrease in both graph size and memory footprint.

Our memory cost analysis demonstrates the significant efficiency gains achieved after compressing $G$ into $G_c$ using SPGC. Through the three key evaluations - Graph Memory (GM), Memoziation Structure Memory (MSM), and Peak Memory (PM), we observe consistent reductions in memory consumption across all three datasets.

**Ablation Analysis**. We next investigate how memoziation structure $\mathcal{T}$, and neighbor recovery affect the inference accuracy and speed-up achieved by SPGC. We also investigated the impact of aggregation method and improvement of model training cost. We present additional tests with details in [1]. We conduct ablation analysis using **Arxiv** to compare SPGC with its two variants: SPGC _w/o_$\mathcal{T}$: SPGC without $\mathcal{T}$, and SPGC _w/o_$\mathcal{T}$_w_1-hop: SPGC without $\mathcal{T}$, but with 1-hop neighbor decompression. We find the following (as shown in Table 9). (1) Incorporating $\mathcal{T}$ into SPGC results in a noteworthy 43.13% increase in inference accuracy, at the cost of a marginal 12.12% reduction in inference speed-up compared to SPGC _w/o_$\mathcal{T}$. This suggests that the memoziation effectively improves inference accuracy while incurring only a small overhead in inference cost. (2) Compared to SPGC _w/o_$\mathcal{T}$, SPGC

| **Arxiv** | \|V\|+\|E\| | GM | MSM | PM |
|---|---|---|---|---|
| $G$ | 1,335,586 | 101.13 MB | - | 1.44 GB |
| $G_c$ | 1,078,756 | 62.44 MB | 3.75 MB | 1.23 GB |
| mcr | 19.23% (ncr) | 38.26% | - | 14.58% |
| **Yelp** | \|V\|+\|E\| | GM | MSM | PM |
| $G$ | 14,671,666 | 1036.03 MB | - | 14.41 GB |
| $G_c$ | 13,711,624 | 812.50 MB | 50.27 MB | 13.24 GB |
| mcr | 6.54% (ncr) | 21.58% | - | 8.09% |
| **Ogbn-Products** | \|V\|+\|E\| | GM | MSM | PM |
| $G$ | 64,308,169 | 1887.47 MB | - | 30.12 GB |
| $G_c$ | 22,211,452 | 536.50 MB | 83.55 MB | 10.49 GB |
| mcr | 65.46% (ncr) | 76.01% | - | 65.19% |

**Table 8: Memory Cost comparison between the original graph $G$ and compressed graph $G_c$ using SPGC. Graph memory (GM) is measured in the PyTorch Geometric Graph Data format, memoziation structure ($\mathcal{T}$) memory (MSM) is measured directly, and peak memory (PM) represents the peak memory usage during inference with a 3-layer GCN (hidden size = 32).**

| | Compression Scheme | GCN | GAT | GraphSAGE |
|---|---|---|---|---|
| Inference Accuracy | SPGC | **0.59** | **0.58** | **0.59** |
| | SPGC_w/o_$\mathcal{T}$ | 0.42 | 0.44 | 0.37 |
| | SPGC_w/o_$\mathcal{T}$_w_1-hop | 0.45 | 0.52 | 0.47 |
| Inference Speed-up | SPGC | 2.27 | 2.42 | 3.94 |
| | SPGC_w/o_$\mathcal{T}$ | **2.57** | **2.85** | **4.24** |
| | SPGC_w/o_$\mathcal{T}$_w_1-hop | 2.18 | 2.23 | 3.54 |

**Table 9: Ablation Studies of Memoziation Structure $\mathcal{T}$ and and Neighbor Decompression w.r.t. Accuracy and Speed-up.**

_w/o_$\mathcal{T}$_w_1-hop demonstrates an average improvement of 16.50% in inference accuracy at a cost of smaller inference speed-up. These verifies the adaptiveness of SPGC in trading inference speed up with model inference accuracy as needed.

## 8.4 Appendix D: Auxiliary Information

**Similarity Merging**. SPGC and its variants leverage both **structural similarity** and **embedding similarity** to construct shared merging. They prioritize structural similarity as defined by our proposed structural equivalence followed by embedding similarity based on input node features for fine-tuning (see Fig. 3.) Since the original graph $G$ may contain nodes with distinct feature vectors, we apply a featurization pre-processing step to discretize input features. This ensures that numerical values within same range are categorized together after featurization, improving consistency in similarity measures. As shown in Table. 8, compared to original $G$, the compressed $G_c$ from SPGC achieves a reduction of the graph size ($|V| + |E|$) by 19.23%, 6.54%, and 65.46% for **Arxiv**, **Yelp**, and **Ogbn-Products** respectively. We define a new metric to directly quantify the percent of similar nodes psn $= 1 - \frac{|V_{dis}|}{|V|}$ in SPGC, where $|V_{dis}|$ is the number of stand-alone nodes in $R^S$ such that they are not merged with any nodes in $V$. For instance, using the equivalence relation defined by SPGC, we found 43.06% nodes in **Arxiv** and 87.42% nodes in ***Ogbn-Products*** are similar by the similarity definition of SPGC. For SPGC, the extent of similarity merging in an arbitrary graph $G$ depends on its inherent structural and embedding similarities. To enhance the performance of compression, our new variant $(\alpha, r)$-SPGC, introduces configurable parameters $\alpha$ and $r$, enabling fine-tuned control over the merging process and improving the compression ratio.